\documentclass[10pt,journal,compsoc]{IEEEtran}

\ifCLASSOPTIONcompsoc
  \usepackage[nocompress]{cite}
\else
  \usepackage{cite}
\fi
\ifCLASSINFOpdf
\else
\fi

\usepackage{graphicx}
\usepackage{amsmath}
\usepackage{multirow}
\usepackage{url}
\newcommand{\RNum}[1]{\uppercase\expandafter{\romannumeral #1\relax}}
\usepackage{amssymb}
\usepackage{UserCommands}

\newcommand{\eg}{\textit{e}.\textit{g}.\ }

\begin{document}
%
\title{Voxel-Mesh Network for Geodesic-Aware 3D Semantic Segmentation of Indoor Scenes}
%
%
%
%

\author{Zeyu Hu,
        Xuyang Bai,
        Jiaxiang Shang,
        Runze Zhang,\\
        Jiayu Dong,
        Xin Wang,
        Guangyuan Sun,
        Hongbo Fu,
        Chiew-Lan Tai
\IEEEcompsocitemizethanks{\IEEEcompsocthanksitem Zeyu Hu, Xuyang Bai, Jiaxiang Shang and Chiew-Lan Tai are with the Hong Kong University of Science and Technology. E-mail: \{zhuam, xbaiad, jshang, taicl\}@cse.ust.hk.
\IEEEcompsocthanksitem Runze Zhang, Jiayu Dong, Xin Wang and Guangyuan Sun are with Tencent Lightspeed \& Quantum Studios. E-mail: \{ryanrzzhang, jiayudong, alexinwang, gerrysun\}@tencent.com.
\IEEEcompsocthanksitem Hongbo Fu is with the School of Creative Media, City University of Hong Kong. E-mail: hongbofu@cityu.edu.hk. {He is the corresponding author of this paper.}
}
}

%
%

\markboth{IEEE TRANSACTIONS ON PATTERN ANALYSIS AND MACHINE INTELLIGENCE, VOL. X, NO. X, MMMMMMM YYYY}%
{Shell \MakeLowercase{\textit{et al.}}: Bare Advanced Demo of IEEEtran.cls for IEEE Computer Society Journals}
%



\IEEEtitleabstractindextext{%
\begin{abstract}
In recent years, sparse voxel-based methods have become the state-of-the-arts for 3D semantic segmentation of indoor scenes, thanks to the powerful 3D CNNs. Nevertheless, being oblivious to the underlying geometry, voxel-based methods suffer from ambiguous features on spatially close objects and struggle with handling complex and irregular geometries due to the lack of geodesic information. In view of this, we present Voxel-Mesh Network (VMNet), a novel 3D deep architecture that operates on the voxel and mesh representations leveraging both the Euclidean and geodesic information. Intuitively, the Euclidean information extracted from voxels can offer contextual cues representing interactions between nearby objects, while the geodesic information extracted from meshes can help separate objects that are spatially close but have disconnected surfaces. To incorporate such information from the two domains, we design an intra-domain attentive module for effective feature aggregation and an inter-domain attentive module for adaptive feature fusion. Experimental results validate the effectiveness of VMNet: specifically, on the challenging ScanNet dataset for large-scale segmentation of indoor scenes, it outperforms the state-of-the-art SparseConvNet and MinkowskiNet (74.6\% vs 72.5\% and 73.6\% in mIoU) with a simpler network structure (17M vs 30M and 38M parameters).
\end{abstract}

}

\maketitle

\IEEEdisplaynontitleabstractindextext

%
\IEEEpeerreviewmaketitle

\ifCLASSOPTIONcompsoc
\IEEEraisesectionheading{\section{Introduction}\label{sec:introduction}}
\else
\section{Introduction}
\label{sec:introduction}
\fi

%
%
%
%

 
\IEEEPARstart{T}{hanks} to the tremendous progress of RGB-D scanning methods in recent years~\cite{whelan2016elasticfusion,kahler2015very,dai2017bundlefusion}, reliable tracking and reconstruction of 3D surfaces using hand-held, consumer-grade devices have {become} possible. Using these methods, large-scale 3D datasets with reconstructed surfaces and semantic annotations are now available~\cite{dai2017scannet,Matterport3D}. Nevertheless, {compared to 3D surface reconstruction, 3D scene understanding, i.e.,} understanding the semantics of {reconstructed} 
scenes, is still a relatively open research problem.

Inspired by the success of 2D CNN in image semantic segmentation~\cite{chen2017deeplab,long2015fully}, researchers have paid much attention to the straightforward extension of this idea {to} 3D, {by performing} volumetric convolution on {regular} 
grids~\cite{maturana2015voxnet,wu20153d,qi2016volumetric}. {Specifically,} surface reconstructions are first projected to a discrete 3D grid representation, and then 3D convolutional filters are applied to extract features by sliding kernels over neighboring {grid} voxels~\cite{song2017semantic,wang2019voxsegnet,zhou2018voxelnet}. {Such features can be smoothly
propagated in the Euclidean domain}
to accumulate strong contextual information. Unfortunately, 
\begin{figure}[htp]
\centering
\includegraphics[width=0.47\textwidth]{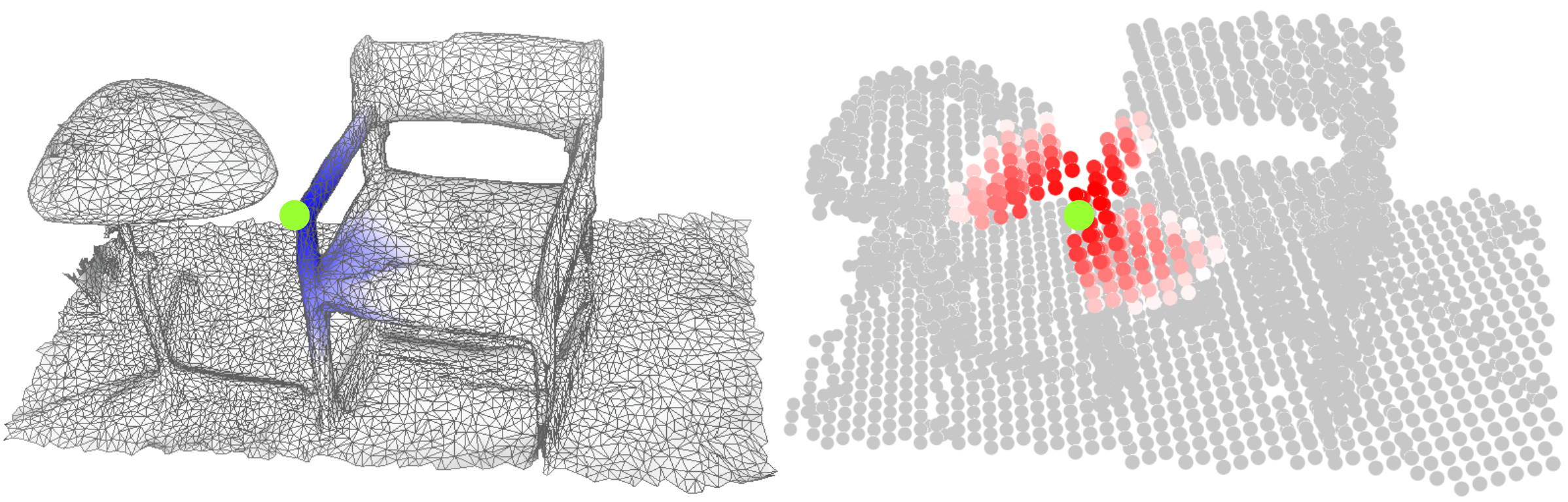}
\caption{\textbf{Illustration of geodesic information loss caused by voxelization.} Considering the green point at the arm of a chair, on {an} 
input 3D mesh surface (\textbf{Left}), its geodesic neighbors (blue) 
can be easily collected, and {the} points of 
different objects are naturally separated. After voxelization (\textbf{Right}), geodesic information is discarded and only Euclidean neighbors (red) that are agnostic to the underlying surface can be extracted. The scan section is taken from the ScanNet dataset~\cite{dai2017scannet}.}
\label{fig:mv_compare}
\end{figure}
dense voxel-based methods require intensive computational power and are thus limited to low-resolution cases~\cite{liu2019point}. To process large-scale data, sparse voxel convolutions~\cite{graham20183d,choy20194d} {have been} 
proposed to lower {the} computational requirement by ignoring inactive voxels. Benefiting from the efficient sparse voxel convolutions, complex networks have
been built, achieving leading results on several 3D semantic segmentation benchmarks~\cite{dai2017scannet,Matterport3D} and outperforming other methods by large margins.

\begin{figure*}[t]
	\centering
	\includegraphics[width=\linewidth]{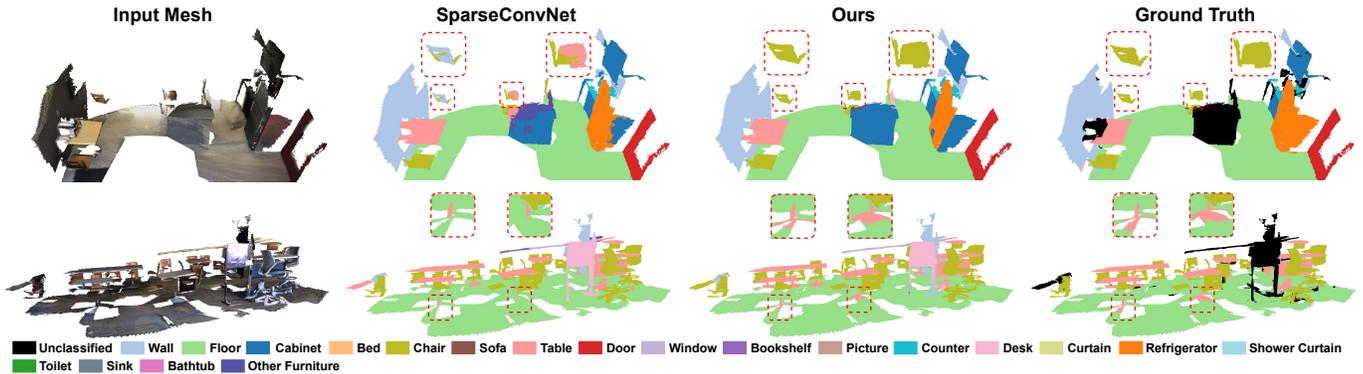}
	\caption{
	\textbf{Limitations of voxel-based methods.} (\textbf{Upper}) Some points of ``chairs'' are mistakenly classified into nearby classes in the Euclidean space by SparseConvNet \cite{graham20183d} since the convolutional filters produce 
	{ambiguous}
	features for spatially close objects. (\textbf{Lower}) On areas with complex and irregular geometries (\eg, the base parts of ``tables''), SparseConvNet fails to predict correct results due to
	{the lack of geodesic information about shape surfaces.}
	}
	\label{fig:teaser}
\end{figure*}

Despite the remarkable achievements, voxel-based methods are not perfect. One of their major limitations is the geodesic information loss caused by the voxelization process (see Fig.~\ref{fig:mv_compare}). 
{Recent public datasets like ScanNet~\cite{dai2017scannet} provide 3D scene reconstructions in the form of high-quality triangular meshes, in which the surface information is naturally encoded.}
On these meshes, {vertices belonging to different objects are well separated, and} geodesic features can be easily aggregated through edge connectivities. 
However, the voxelization process omits all mesh edges and only retains Euclidean positions of mesh vertices. 
Consequently, convolutional filters operating on voxels are agnostic to the underlying surfaces and, therefore, result in two problems. First, these filters generate similar features for voxels that are close in the Euclidean domain, even though these voxels may belong to different objects and are distant in the geodesic domain. As shown in the top example of Fig.~\ref{fig:teaser}, these 
{ambiguous}
features 
produce sub-optimal predictions for objects that are spatially close. 
{Second, without the geodesic information about shape surfaces, these Euclidean convolutions may struggle with learning specific object shapes.}
As shown in the lower example of Fig.~\ref{fig:teaser}, this property is
problematic for segmentation on areas with complex and irregular geometries.

We have discussed the advantages of voxel-based methods on contextual learning and their problems on geodesic information loss. It is appealing to design a method resolving the problems while retaining the{se} advantages by leveraging both the Euclidean and geodesic information. A 
{possible solution} 
is to take voxels and the original meshes as the sources for {the} Euclidean and geodesic information, respectively. It is therefore natural to ask how these two representations can be combined in a common architecture. 

{To address this question}, we propose the Voxel-Mesh network (VMNet), a novel deep hierarchical architecture for geodesic-aware 3D semantic segmentation. Starting from a mesh representation, to extract informative contextual features in the Euclidean domain, we first voxelize the input mesh and apply sparse voxel convolutions. Next, to incorporate {the} geodesic information, the extracted contextual features are projected from the Euclidean domain to the geodesic domain, specifically, from voxels to mesh vertices. These projected features are further fused and aggregated {to combine} 
both the Euclidean and geodesic information. 

In order to build such a deep architecture that is capable of {effectively} learning useful features incorporating information from the two domains, it is critical to design proper ways to aggregate intra-domain features and to fuse inter-domain features. In view of the great success of self-attention operators for feature processing~\cite{NIPS2017_3f5ee243,parmar2018image,li2020neural}, we therefore present two key components of VMNet: Intra-domain Attentive Aggregation Module and Inter-domain Attentive Fusion Module. The former aims to aggregate the projected features on the original meshes to incorporate {the} geodesic information and the latter focuses on the effective fusion of features from {the} two domains.

\begin{figure*}[t]
	\centering
	\includegraphics[width=\linewidth]{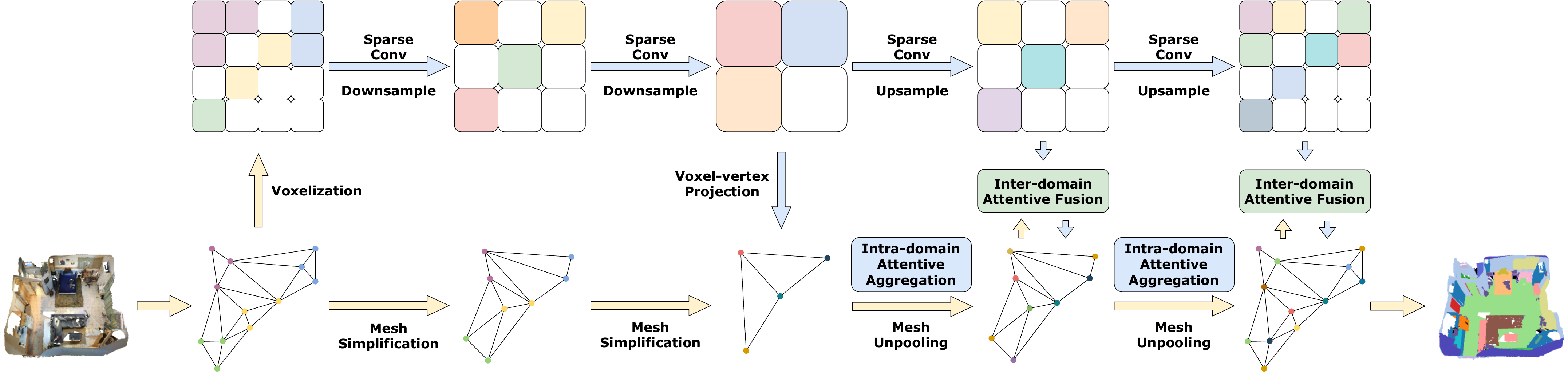}
	\caption{
	\textbf{Overview of Voxel-Mesh Network (VMNet).} Taking a colored mesh as input, we first rasterize it and apply voxel-based sparse convolutions to extract contextual information in the Euclidean domain. These features are then projected from voxels to vertices, and are further aggregated and fused in the geodesic domain producing distinctive per-vertex features. {For simplicity, skip connections {between the encoder and decoder} 
	are neglected here and only three levels of hierarchical voxel downsampling and mesh simplification are shown. 
	}
	}
	\label{fig:overview}
\end{figure*}

We conduct extensive experiments to demonstrate the effectiveness of our method on the popular ScanNet v2 benchmark~\cite{dai2017scannet} and the recent Matterport3D benchmark~\cite{Matterport3D}. {VMNet outperforms existing sparse voxel-based methods SparseConvNet~\cite{graham20183d} and MinkowskiNet~\cite{choy20194d} (74.6\% vs 72.5\% and 73.6\% in mIoU) with {a} simpler network structure (17M vs 30M and 38M parameters) on the ScanNet dataset and set{s} {a} 
new state-of-the-art
on the Matterport3D dataset.} 

To summarize, our contributions are threefold:
\begin{enumerate}
\item We propose a novel deep architecture, VMNet, {which} 
operates on the voxel and mesh representations, leveraging both the Euclidean and geodesic information.
\item We propose an intra-domain attentive aggregation module, which effectively refines geodesic features through edge connectivities.
\item We propose an inter-domain attentive fusion module, which adaptively combines Euclidean and geodesic features.
\end{enumerate}

\section{Related Work} \label{related}

{
In this section, we first {review} 
relevant works on 3D semantic segmentation, organized according to their inherent convolutional categories, and then discuss the application of attention mechanism in 3D semantic segmentation.

\noindent\textbf{2D-3D.}
A conventional way of performing 3D semantic segmentation is to first represent 3D shapes through their 2D projections from various viewpoints, and then leverage existing {image segmentation} techniques and architectures from the 2D domain~\cite{lawin2017deep,kundu2020virtual}. Instead of choosing a global projection viewpoint, some researchers have proposed {to} 
project local neighborhoods to local tangent planes and process them with 2D convolutions~\cite{tatarchenko2018tangent,yang2020pfcnn,huang2019texturenet}. Taking the RGB frames as additional inputs, other researchers have proposed methods that combine 2D and 3D features through 2D-3D projection~{\cite{dai20183dmv,BPNet}}. Although these methods can easily benefit from the success of {image segmentation techniques (mainly based on 2D CNNs)}, 
they often require a large amount of additional 2D data, involve {a} complex multi-view projection process, and rely heavily on viewpoint selection. {Some of these methods have attempted to} 
utilize geodesic information implicitly through mesh textures~\cite{huang2019texturenet} or point normal~\cite{tatarchenko2018tangent}. They achieve fairly decent results
but fail to fully exploit the geodesic information.

\noindent\textbf{PointConv \& SparseConv.}
Partly due to the difficulties of handling mesh edges in deep neural networks, most existing 3D semantic segmentation methods take raw point clouds or transformed voxels as input~{\cite{boulch2017unstructured,lawin2017deep,roynard2018classification,ben20183dmfv,riegler2017octnet,qi2017pointnet,qi2017pointnet++,nekrasov2021mix3d,wang2017cnn}}. Point-based methods apply convolutional kernels to the local neighborhoods of points obtained using k-NN or spherical search~{\cite{zhao2019pointweb,wang2019dynamic,wang2019graph,su2018splatnet,hua2018pointwise,wu2019pointconv,hu2020jsenet}}.
Numerous designs of point-based convolutional kernels have been proposed~\cite{lei2019octree,komarichev2019cnn,thomas2019kpconv,mao2019interpolated,zhang2018efficient}. In the case of voxel-based methods, the raw 3D data is first transformed into a voxel representation and then {processed by} standard CNNs
~\cite{maturana2015voxnet,qi2016volumetric,wang2019voxsegnet,zhou2018voxelnet,huang2021supervoxel}. To address the cubic memory and computation consumption problem of voxel-based operations, recent works have made efforts to propose efficient sparse voxel convolutions~\cite{graham20183d,choy20194d,tang2020searching}. In both point-based and voxel-based methods, features are aggregated over the Euclidean space only. In contrast, we additionally consider geodesic information of the underlying object surfaces.

\noindent\textbf{GraphConv.}
Graph convolution networks can be grouped into spectral networks~\cite{defferrard2016convolutional,simonovsky2017dynamic} and local filtering networks~\cite{masci2015geodesic,boscaini2016learning,monti2017geometric}. Spectral networks work well on clean synthetic data, but are sensitive to reconstruction noise and are thus not 
applicable 
to 3D semantic segmentation. Local filtering networks define handcrafted coordinate systems and apply convolutional operations over patches. For 3D semantic segmentation, these methods often perform over local neighborhoods of point clouds~\cite{jiang2019hierarchical,lei2020spherical} and are thus oblivious to the underlying geometry. 

Our method falls into both the SparseConv and GraphConv categories. It is similar in spirit to the recent work of Schult et al.~\cite{schult2020dualconvmesh}, which combines a Euclidean-based and a geodesic-based graph convolutions. However, instead of concatenating features obtained from different convolutional filters as done in \cite{schult2020dualconvmesh}, we first accumulate strong contextual information in the Euclidean domain and then adaptively fuse and aggregate geometric information in the geodesic domain, leading to a significant better segmentation performance (see Section~\ref{Results}).

\noindent\textbf{Attention.}
For 3D semantic segmentation, most existing methods implement attention layers operating on the local neighborhoods of point clouds for feature aggregation~\cite{fuchs2020se3transformers,wang2019graph} or on downsampled point sets for context augmentation~\cite{yan2020pointasnl,Wong_2020}. In our work, instead of operating on point clouds, we build attentive operators applying on triangular meshes. Moreover, in contrast to previous works that process features {in a single} 
domain, we propose both an intra-domain module and an inter-domain module.

}

\section{METHODOLOGY} \label{method}


{In this section, we first {give an} 
overview of our voxel-mesh network in Section~\ref{sec::overview}. Then we introduce the network architecture in Section~\ref{network_arc}. The voxel-based contextual feature aggregation branch is described in Section~\ref{voxel}. Section{s}~\ref{intra} and \ref{inter} depict the proposed attentive modules for intra-domain feature aggregation and inter-domain feature fusion{, respectively}. 
Finally, we discuss two well-known mesh simplification methods{, which are used to} 
build a mesh hierarchy for multi-level feature learning
in Section~\ref{mesh}.}

{\subsection{Overview} \label{sec::overview}} 
VMNet deals with two types of 3D representations: voxels and meshes. As depicted in Fig.~\ref{fig:overview}, {the network consists of two branches: according to their operating domains, we denote the upper one as the \textit{Euclidean branch} and the lower one as the \textit{geodesic branch}.}

To accumulate contextual information in the Euclidean domain, {taking a mesh as input, the colored vertices}
are first voxelized and then fed to the Euclidean branch. Building on sparse voxel-based convolutions, we construct a U-Net~\cite{ronneberger2015u} like encoder-decoder structure, where the encoder is symmetric to the decoder, including skip connections between both. Multi-level sparse voxel-based feature maps $(\mathcal{S}^0,...,\mathcal{S}^l,...,\mathcal{S}^L)$ can be extracted from the decoder.

\begin{figure*}[t]
	\centering
	\includegraphics[width=\linewidth]{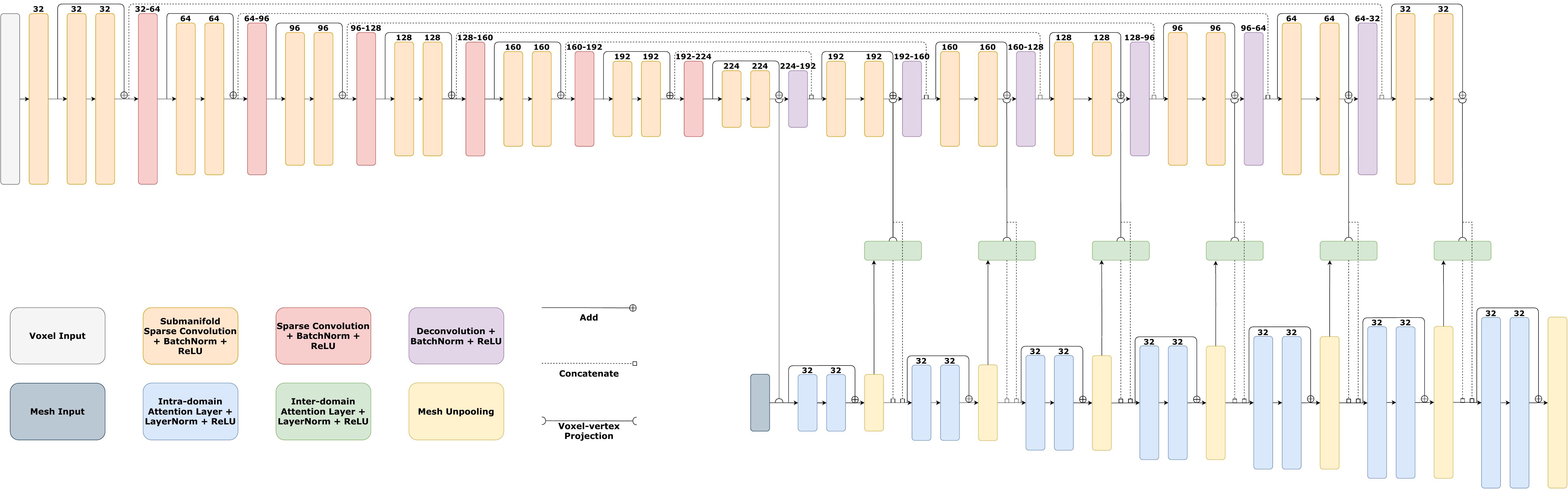}
	\caption{
	{\textbf{Network architecture of VMNet.} We use a voxel-based 3D U-Net~\cite{ronneberger2015u} as the contextual feature extractor{, consisting of an encoder and a decoder}. Afterwards, at each level of the decoder, the aggregated contextual features are first projected from the Euclidean domain to the geodesic domain, and then processed by the intra-domain attentive aggregation modules and the inter-domain attentive fusion modules defined over triangular meshes, yielding distinctive per-vertex features enriched with both the Euclidean and geodesic information. The number above each layer indicates {the number of its corresponding feature channel.} 
	}
	}
	\label{fig:vmnet_detailed}
\end{figure*}

\begin{figure}[t]
\centering
\includegraphics[width=0.44\textwidth]{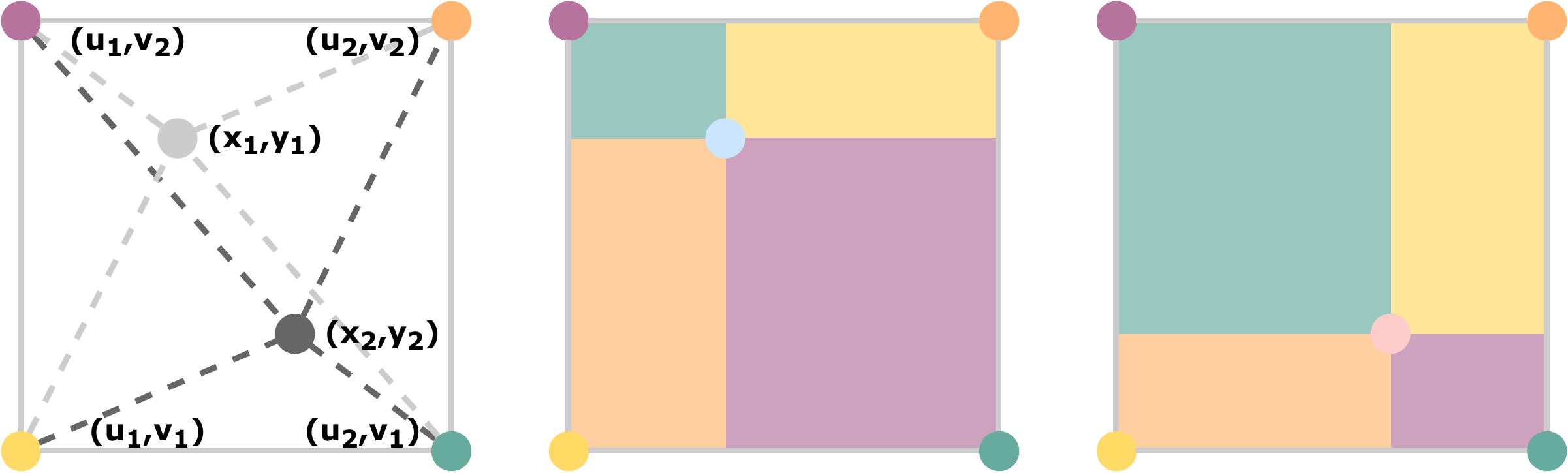}
\caption{\textbf{2D illustration of voxel-vertex projection.} Vertices $(x_1,y_1)$ {and} 
$(x_2,y_2)$ share the same set of neighboring voxels but
their projected features are different
through trilinear interpolation {(bilinear interpolation for the 2D case)}.}
\label{fig:trilinear}
\end{figure}

Although these contextual features offer valuable semantic cues for scene understanding, their unawareness of the underlying geometric surfaces will lead to sub-optimal results. Therefore, to incorporate geodesic information, the accumulated contextual features are projected from the Euclidean domain to the geodesic domain for further processing {(Section~\ref{voxel})}. {In the geodesic branch, we prepare a hierarchy of simplified meshes $(\mathcal{M}^0,...,\mathcal{M}^l,...,\mathcal{M}^L)$, in which each level of simplified mesh $\mathcal{M}^l$ corresponds to a downsampling level of sparse voxels $\mathcal{S}^l$. Trace maps of the mesh simplification processes are saved for unpooling operations between mesh levels.}
{At the first level of the decoding process (level $L$), the features are projected from voxels $\mathcal{S}^L$ to mesh vertices $\mathcal{M}^L$ and then refined through intra-domain attentive aggregation {(Section~\ref{intra})}. The resulting geodesic features of $\mathcal{M}^L$ are unpooled to the next level $\mathcal{M}^{L-1}$. At each following level $l$, the Euclidean features projected from $\mathcal{S}^l$ and the unpooled geodesic features of $\mathcal{M}^l$ are first adaptively combined through inter-domain attentive fusion {(Section~\ref{inter})} and then the fused features are further refined through intra-domain attentive aggregation before being unpooled to the next level.}

{
\subsection{Network Architecture} \label{network_arc}
The network architecture adopted in VMNet is {illustrated} in Fig.~\ref{fig:vmnet_detailed}. In the upper branch (i.e., the Euclidean branch), the network is mainly built upon submanifold sparse convolution (SSC) layers and sparse convolution (SC) layers, both of which are originally introduced by Graham {et al.}~\cite{graham20183d} 
for 3D semantic segmentation.
The basic building module is a residual block consisting of two layers of SSC with a skip connection implemented by addition. Each pair of adjacent residual blocks {are} 
connected through an SC layer{, which} 
performs downsampling (encoding stage) or upsampling (decoding stage)
of the sparse voxels $\mathcal{S}$. {In total,} there are 13 residual blocks and 7 levels of sparse voxels $(\mathcal{S}^0,...,\mathcal{S}^l,...,\mathcal{S}^6)$. In the lower branch (i.e., the geodesic branch), similar to the upper one, the network is constructed by residual blocks, each consisting of two layers of our proposed intra-domain attention layers (Section~\ref{intra}){,} which operate on the triangular meshes $\mathcal{M}$. Each pair of adjacent residual blocks in the geodesic branch {are} 
connected through a mesh unpooling layer. The distinctive per-vertex features on the last mesh level $\mathcal{M}^0$ are used for semantic prediction. Between the two branches, the projected Euclidean features and the aggregated geodesic features are adaptively fused through our proposed inter-domain attention layers (Section~\ref{inter}).
}

\subsection{Voxel-based Contextual Feature Aggregation}\label{voxel}

\noindent\textbf{Voxelization.} 
At mesh level $\mathcal{M}^0$, with all edge connectivities omitted, the input features {(colors)} of mesh vertices {$\{(V_i, f_i)\}$ }
are transformed into the voxel cells $\{V_{u,v,w}\}$ by averaging all features $f_i$ whose {corresponding} coordinate $V_i:(x_i,y_i,z_i)$ falls into the voxel cell $(u,v,w)$: 
    \begin{equation}
    \begin{split}
        f_{u,v,w} = \frac{1}{N_{u,v,w}}\sum_{i=1}^nB[floor(x_i \cdot r)=u, \\
        floor(y_i \cdot r)=v,floor(z_i \cdot r)=w] \cdot f_i,
        \label{equation:voxelization}
    \end{split}
    \end{equation} 
where $r$ denotes the voxel resolution, $B[\cdot]$ is the binary indicator of whether vertex $V_i$ belongs to the voxel cell $(u,v,w)$, and $N_{u,v,w}$ is the number of vertices falling into that cell~\cite{liu2019point}. 

\noindent\textbf{Contextual Feature Aggregation.} 
To accumulate contextual information, we construct a simple U-Net~\cite{ronneberger2015u} structure based on voxel convolutions. We adopt the sparse implementation provided by \cite{tang2020searching}.

\noindent\textbf{Voxel-vertex
Projection.} 
With the contextual features aggregated in the Euclidean domain,
{at each level $l$, we transform the features of voxels $\mathcal{S}^l$ back to vertices $\mathcal{M}^l$ for further processing in the geodesic domain. Inspired by previous works~\cite{liu2019point,tang2020searching},
we compute each vertex's feature utilizing trilinear interpolation over its neighboring eight voxels.
Through this means, the projected features are distinct even for {the} vertices sharing the same set of neighboring voxels. A 2D illustration of the projection is {shown} in Fig.~\ref{fig:trilinear}.
}

\begin{figure}[t]
\centering
\includegraphics[width=0.47\textwidth]{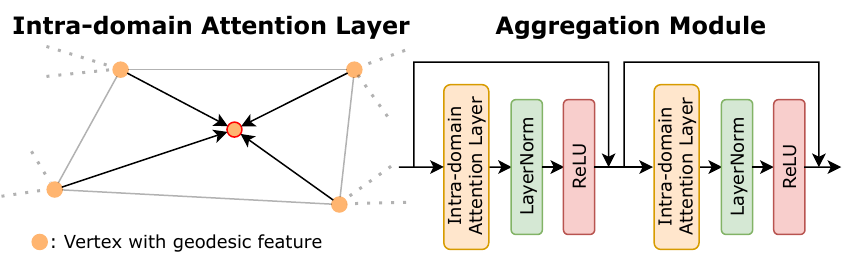}
\caption{\textbf{Illustration of intra-domain attentive aggregation module.} \textbf{(Left)} Intra-domain attention layer operates on mesh vertices aggregating geodesic information through edge connectivities. \textbf{(Right)} The aggregation module consists of two attention layers with skip connections.}
\label{fig:intra}
\end{figure}

\subsection{Intra-domain Attentive Aggregation Module} \label{intra}
After contextual feature aggregation and voxel-vertex projection, to effectively refine the projected features, we design an intra-domain attentive aggregation module operating on the geodesic domain. As shown in Fig.~\ref{fig:intra} {(Left)}, at each mesh level, we perform attentive aggregation on the graph $\mathcal{G}=(\mathcal{V},E)$ induced by the underlying mesh $\mathcal{M}$. Note that we neglect the level superscript $l$ to ease readability. Our intra-domain attention layer is based on the standard scalar attention~\cite{NIPS2017_3f5ee243}{, which is often used for point clouds in 3D semantic segmentation, but not for triangular meshes.}
Specifically, {at layer $k$,} the output feature $f_i^{'geo}$ of vertex $V_i$ with an input feature $f_i^{geo}$ is computed as:
    \begin{equation}
    \begin{split}
        f_i^{'geo} = \rho^{intra}_k(f_i^{geo}) + \sum_{j \in N_i}\omega_{ij}\alpha^{intra}_k(f_j^{geo}), \\
        \omega_{ij} = softmax(\frac{\varphi^{intra}_k(f_i^{geo})^T\psi^{intra}_k(f_j^{geo})}{\sqrt{d}}),
        \label{equation:intra}
    \end{split}
    \end{equation} 
where $N_i$ is the one-ring neighborhood of vertex $V_i$. 
{The functions} {$\rho^{intra}_k$, $\alpha^{intra}_k$, $\varphi^{intra}_k$, and $\psi^{intra}_k$ are vertex-wise feature transformations implemented by MLP}, $\omega_{ij}$ is the attention coefficient, and $d$ is the size of output feature channels. Since the positional information is naturally embedded in the voxel-based contextual feature aggregation step, we do not implement a position encoding function explicitly. Our attention layer is inspired by the implementation in \cite{shi2020masked}, which operates on abstract graphs for semi-supervised node classification, while {our method}
operates on 3D mesh graphs for geodesic feature aggregation.

Building on the intra-domain attention layer, we design an aggregation module performing two steps of attentive feature aggregation on each simplified mesh level (see Fig.~\ref{fig:intra} {(Right)}).

\subsection{Inter-domain Attentive Fusion Module} \label{inter}

\begin{figure}[t]
\centering
\includegraphics[width=0.47\textwidth]{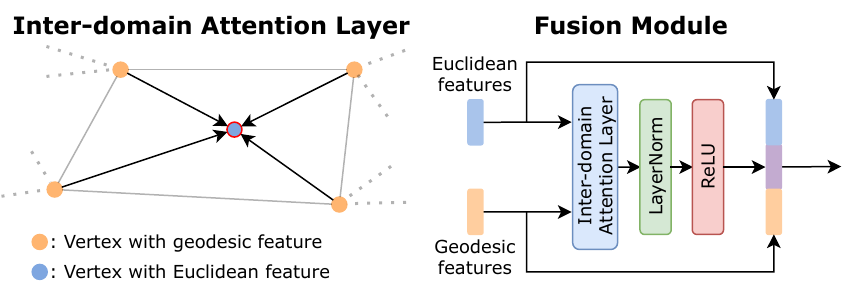}
\caption{\textbf{Illustration of inter-domain attentive fusion module.} \textbf{(Left)} 
{Inter-domain attention layer adaptively combines geodesic features and Euclidean features on mesh vertices.}
\textbf{(Right)} The fused feature map generated by the inter-domain attention layer is further combined with the original geodesic feature map and the projected Euclidean feature map through concatenation.}
\label{fig:inter}
\end{figure}

{Operating} 
on both the voxel and mesh representations {poses} 
a demand for Euclidean and geodesic feature fusion. To adaptively combine features from {the} two domains, we propose an inter-domain attentive fusion module. As depicted in Fig.~\ref{fig:inter} (Left), between each pair of sparse voxel level $\mathcal{S}$ and mesh level $\mathcal{M}$ (except for level $L$), we perform attentive fusion on the same graph $\mathcal{G}=(\mathcal{V},E)$ as the one used for intra-domain aggregation (level superscript $l$ is neglected).
However, unlike intra-domain attention{, which} 
processes features in the same domain, inter-domain attention takes as input both the geodesic features $f^{geo}$ and the Euclidean features $f^{euc}$ projected from voxels. At layer $k$, the fused feature $f_i^{fuse}$ of vertex $V_i$ is computed as:
    \begin{equation}
    \begin{split}
        f_i^{fuse} = \rho^{inter}_k(f_i^{euc}) + \sum_{j \in N_i}\omega_{ij}\alpha^{inter}_k(f_j^{geo}), \\
        \omega_{ij} = softmax(\frac{\varphi^{inter}_k(f_i^{euc})^T\psi^{inter}_k(f_j^{geo})}{\sqrt{d}}),
        \label{equation:inter}
    \end{split}
    \end{equation} 
where $N_i$ is the same one-ring neighborhood of vertex $V_i$ as the one used for intra-domain aggregation. Unlike the one in intra-domain attention, the inter-domain attention coefficient $\omega_{ij}$ is conditioned on both the Euclidean and geodesic features enabling the network to adaptively fuse features from {the} two domains.


As shown in Fig.~\ref{fig:inter} {(Right)}, the proposed inter-domain attentive fusion module takes both the Euclidean features and the geodesic features as inputs. These features are fed to one inter-domain attention layer followed by 
{layer normalization} and {ReLU} 
activation. Before being passed {on} for further processing, the fused feature map is concatenated with the projected Euclidean feature map and the original geodesic feature map.

\subsection{Mesh Simplification} \label{mesh}
{
To construct a deep architecture for multi-level feature learning, we generate a hierarchy of mesh levels $(\mathcal{M}^0,...,\mathcal{M}^l,...,\mathcal{M}^L)$ of increasing simplicity{,} interlinked by pooling trace maps. Each level of simplified mesh corresponds to a level of downsampled 3D sparse voxels. For mesh simplification, there are two 
well-known
methods from the geometry processing domain: Vertex Clustering (VC)~\cite{rossignac1993multi} and Quadric Error Metrics (QEM)~\cite{garland1997surface}. 
{As shown in Fig.~\ref{fig:vc},} 
{d}uring the vertex clustering process, a 3D uniform grid with cubical cells of a fixed side length is placed over the input graph and all vertices falling into the same cell are grouped. This generates uniform-sampled simplified meshes, {possibly} with topology changes and non-manifold faces. On the contrary, {the} QEM method incrementally collapses mesh edges according to an approximate error of the geometric distortion introduced by this collapse, and thus has explicit control over mesh topology
{(see Fig.~\ref{fig:qem}).}
Since our goal is to extract meaningful geodesic information, we prefer {the} QEM method for its better topology-preserving property.
\begin{figure}[htp]
\centering
\includegraphics[width=0.45\textwidth]{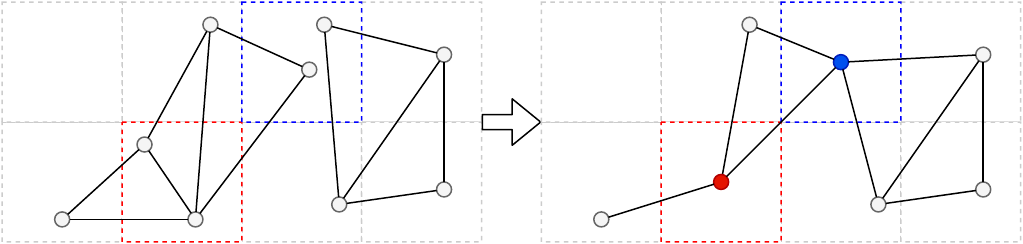}
\caption{{\textbf{Illustration of Vertex Clustering for mesh simplification.} Vertices falling in the same cell are merged to form a new vertex. The resulting mesh {might} 
be non-manifold (\textbf{red cell}) {or have its topology changed} 
(\textbf{blue cell}).}}
\label{fig:vc}
\end{figure}
\begin{figure}[htp]
\centering
\includegraphics[width=0.43\textwidth]{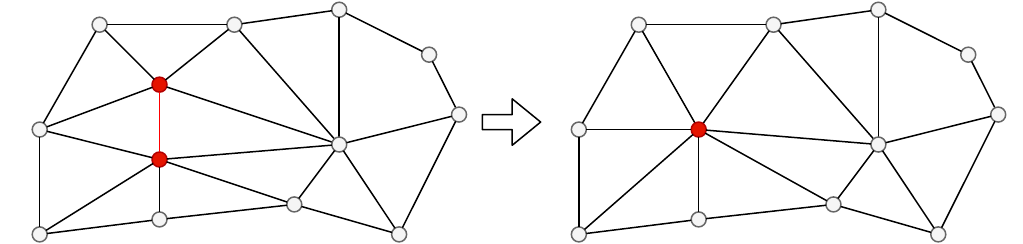}
\caption{{\textbf{Illustration of Quadric Error Metrics based edge collapse for mesh simplification.} The edge between two red vertices is collapsed and the resulting mesh is re-triangulated with {its} topology preserved.}}
\label{fig:qem}
\end{figure}
{However, directly applying the QEM method on the 
original meshes 
results in high-frequency signals in noisy areas~\cite{schult2020dualconvmesh}. Therefore, we apply the VC method on the original mesh for the first two mesh levels and then apply the QEM method for the remaining mesh levels.} 
{We present an ablation study on mesh simplification methods in Section~\ref{Ablation}.}

}

\section{EXPERIMENTS}
To demonstrate the effectiveness of our proposed method, we now present various experiments conducted on two large-scale 3D scene segmentation datasets, which contain meshed point clouds of various indoor scenes. We first introduce the datasets and evaluation metrics {that we used} in Section~\ref{Datasets}, and then present the implementation details for reproduction in Section~\ref{Implementation}. We report the results on the ScanNet and Matterport3D datasets in Section~\ref{Results}, and the ablation studies in Section~\ref{Ablation}.

\subsection{Datasets and Metrics} \label{Datasets}

\noindent\textbf{ScanNet v2~\cite{dai2017scannet}.}
ScanNet dataset contains 3D meshed point clouds of a wide variety of indoor scenes. 
{Each scene is provided with semantic annotations and reconstructed surfaces represented by a textured mesh.}
The dataset contains 20 valid semantic classes. We perform all our experiments using the public training, validation, and test split of 1201, 312, and 100 scans, respectively. 

\noindent\textbf{Matterport3D~\cite{Matterport3D}.}
Matterport3D is a large RGB-D dataset of 90 building-scale scenes. Similar to ScanNet, the full 3D mesh reconstruction of each building and semantic annotations are provided. The dataset contains 21 valid semantic classes. Following previous works~\cite{schult2020dualconvmesh,qi2017pointnet++,su2018splatnet,tatarchenko2018tangent,dai20183dmv,huang2019texturenet}, we split the whole dataset into training, validation, and test sets of size 61, 11, and 18, respectively.

\noindent\textbf{Metrics.}
For evaluation, we use the same protocol as introduced in previous works~\cite{schult2020dualconvmesh,qi2017pointnet++,choy20194d,graham20183d}. We report mean class intersection over union (mIoU) results for ScanNet and mean class accuracy for Matterport3D. During testing, we project the semantic labels to the vertices of the original meshes and test directly on meshes.

\begin{table}
\caption[Caption for LOC]{\textbf{Mean intersection over union scores on ScanNet Test~\cite{dai2017scannet}.} Detailed results can be found on the ScanNet benchmarking website\footnotemark and in the supplementary material. 
}
\label{table::scannet}
\begin{center}
\begin{tabular}{r|c|l}
\hline
Method & mIoU(\%) & Conv Category \\
\hline\hline
TangentConv~\cite{tatarchenko2018tangent} & 43.8 & \multirow{8}{3cm}{2D-3D} \\
SurfaceConvPF~\cite{yang2020pfcnn} & 44.2 & \\
3DMV~\cite{dai20183dmv} & 48.3 & \\
TextureNet~\cite{huang2019texturenet} & 56.6 & \\
JPBNet~\cite{chiang2019unified} & 63.4 & \\
MVPNet~\cite{jaritz2019multi} & 64.1 & \\
V-MVFusion~\cite{kundu2020virtual} & 74.6 & \\
BPNet~\cite{BPNet} & \textbf{74.9} & \\
\hline
PointNet++~\cite{qi2017pointnet++} & 33.9 & \multirow{8}{3cm}{PointConv} \\
FCPN~\cite{rethage2018fully}    & 44.7 & \\
PointCNN~\cite{NEURIPS2018_f5f8590c} & 45.8 & \\
DPC~\cite{engelmann2020dilated} & 59.2 & \\
MCCN~\cite{hermosilla2018monte} & 63.3 & \\
PointConv~\cite{wu2019pointconv} & 66.6 & \\
KPConv~\cite{thomas2019kpconv} & 68.4 & \\
JSENet~\cite{hu2020jsenet} & 69.9 & \\
\hline
SparseConvNet~\cite{graham20183d} & 72.5 & \multirow{2}{3cm}{SparseConv} \\
MinkowskiNet~\cite{choy20194d} & 73.6 & \\
\hline
SPH3D-GCN~\cite{lei2020spherical} & 61.0 & \multirow{3}{3cm}{GraphConv} \\
HPEIN~\cite{jiang2019hierarchical} & 61.8 & \\
DCM-Net~\cite{schult2020dualconvmesh} & 65.8 & \\
\hline\hline
VMNet (\textbf{Ours}) & \textbf{74.6} & Sparse+Graph Conv \\
\hline
\end{tabular}
\end{center}
\end{table}
\footnotetext{\url{http://kaldir.vc.in.tum.de/scannet_benchmark/}}

\subsection{Implementation Details} \label{Implementation}

In this section, we discuss the implementation details for our experiments. VMNet is coded in Python and PyTorch (Geometric)~\cite{Fey/Lenssen/2019,paszke2017automatic}. All the experiments are conducted on one NVIDIA Tesla V100 GPU.

\noindent\textbf{Data Preparation.}
{We perform training and inference on full meshes without cropping. For the Euclidean branch of VMNet, input meshes are voxelized at a resolution of 2 cm. To compute the hierarchical mesh levels accordingly for the geodesic branch, we first apply the VC method on the input mesh with {the respective} cubical cell lengths of 2 cm and 4 cm for the first two mesh levels. For each remaining level, the QEM method is applied to simplify the mesh until the vertex number is reduced to 30\% of its preceding mesh level. For better generalization ability, edges of all mesh levels are randomly sampled during training. We use the vertex colors as the only input features and apply data augmentation, including random scaling, rotation around the gravity axis, spatial translation, and chromatic jitter.}

\noindent\textbf{Training Details.}
We train the network end-to-end by minimizing the cross entropy loss using Momentum SGD with the Poly scheduler {decaying} from 
learning rate 1e-1.

\begin{table*}[th]
    \caption{\textbf{Mean class accuracy scores on the Matterport3D Test~\cite{Matterport3D}.} The same network definition as for the ScanNet benchmark is used. Conv Category: (\RNum{1}) 2D-3D, (\RNum{2}) PointConv, (\RNum{3}) VoxelConv, (\RNum{4}) GraphConv, (\RNum{5}) Sparse+Graph Conv.}
    \label{table:matterport}    
    \centering
    \resizebox{\textwidth}{!}{
        \begin{tabular}{r|c|c|ccccccccccccccccccccc}
        \hline
        Method & mAcc(\%) & Cat & wall & floor & cab & bed & chair & sofa & table & door & wind & shf & pic & cntr & desk & curt & ceil & fridg & show & toil & sink & bath & other \\ 
        \hline
        TangentConv~\cite{tatarchenko2018tangent} & 46.8 & \RNum{1} & 56.0 & 87.7 & 41.5 & 73.6 & 60.7 & 69.3 & 38.1 & 55.0 & 30.7 & 33.9 & 50.6 & 38.5 & 19.7 & 48.0 & 45.1 & 22.6 & 35.9 & 50.7 & 49.3 & 56.4 & 16.6 \\
        3DMV~\cite{dai20183dmv} & 56.1 & \RNum{1} & 79.6 & 95.5 & 59.7 & 82.3 & 70.5 & 73.3 & 48.5 & 64.3 & 55.7 & 8.3 & 55.4 & 34.8 & 2.4 & \textbf{80.1} & 94.8 & 4.7 & 54.0 & 71.1 & 47.5 & 76.7 & 19.9 \\
        TextureNet~\cite{huang2019texturenet} & 63.0 & \RNum{1} & 63.6 & 91.3 & 47.6 & 82.4 & 66.5 & 64.5 & 45.5 & 69.4 & 60.9 & 30.5 & \textbf{77.0} & \textbf{42.3} & 44.3 & 75.2 & 92.3 & 49.1 & 66.0 & 80.1 & \textbf{60.6} & 86.4 & 27.5 \\
        SplatNet~\cite{su2018splatnet} & 26.7 & \RNum{2} & \textbf{90.8} & 95.7 & 30.3 & 19.9 & 77.6 & 36.9 & 19.8 & 33.6 & 15.8 & 15.7 & 0.0 & 0.0 & 0.0 & 12.3 & 75.7 & 0.0 & 0.0 & 10.6 & 4.1 & 20.3 & 1.7 \\
        PointNet++~\cite{qi2017pointnet++} & 43.8 & \RNum{2} & 80.1 & 81.3 & 34.1 & 71.8 & 59.7 & 63.5 & \textbf{58.1} & 49.6 & 28.7 & 1.1 & 34.3 & 10.1 & 0.0 & 68.8 & 79.3 & 0.0 & 29.0 & 70.4 & 29.4 & 62.1 & 8.5 \\
        ScanComplete~\cite{dai2018scancomplete} & 44.9 & \RNum{3} & 79.0 & \textbf{95.9} & 31.9 & 70.4 & 68.7 & 41.4 & 35.1 & 32.0 & 37.5 & 17.5 & 27.0 & 37.2 & 11.8 & 50.4 & \textbf{97.6} & 0.1 & 15.7 & 74.9 & 44.4 & 53.5 & 21.8 \\ 
        DCM-Net~\cite{schult2020dualconvmesh} & 66.2 & \RNum{4} & 78.4 & 93.6 & \textbf{64.5} & \textbf{89.5} & 70.0 & \textbf{85.3} & 46.1 & \textbf{81.3} & \textbf{63.4} & 43.7 & 73.2 & 39.9 & \textbf{47.9} & 60.3 & 89.3 & \textbf{65.8} & 43.7 & \textbf{86.0} & 49.6 & 87.5 & \textbf{31.1} \\
        \hline
        VMNet (\textbf{Ours}) & \textbf{67.2} & \RNum{5} & 85.9 & 94.4 & 56.2 & \textbf{89.5} & \textbf{83.7} & 70.0 & 54.0 & 76.7 & 63.2 & \textbf{44.6} & 72.1 & 29.1 & 38.4 & 79.7 & 94.5 & 47.6 & \textbf{80.1} & 85.0 & 49.2 & \textbf{88.0} & 29.0 \\
        \hline
 		\end{tabular}
       }

\end{table*}

\begin{table}
    \caption{\textbf{Comparison of run-time complexity against SOTA sparse voxel-based methods.} For {a} fair comparison, we report the latencies of both their original versions (Ori) and our implementations using the same type of sparse convolution (TS) as VMNet. 
    }
    \label{table::complexity}
    \centering
    \resizebox{0.47\textwidth}{!}{
    \begin{tabular}{r|c|cc|c}
    \hline
    \multirow{2}{1cm}{Method} & \multirow{2}{1.7cm}{Params (M)} & \multicolumn{2}{c|}{Latency (ms)} & \multirow{2}{1.2cm}{mIoU(\%)} \\
     & & Ori & TS & \\ 
    \hline
    SparseConvNet~\cite{graham20183d} & 30.1 & 712 & 102 & 72.5 \\
    MinkowskiNet~\cite{choy20194d} & 37.8 & 629 & 105 & 73.6 \\
    VMNet (\textbf{Ours}) & 17.5 & - & 107 & \textbf{74.6} \\
    \hline
    \end{tabular}
    }
\end{table}

\begin{table}[h]
    \caption{{\textbf{Comparison of run-time complexity against SOTA methods of different convolutional categories. 
    }}
    }
    \label{table::complexity_sup}
    \centering
    \resizebox{0.47\textwidth}{!}{
    \begin{tabular}{r|c|c|c|c}
    \hline
    {Method} & {Conv Category} & {Params (M)} & {Latency (ms)} & {mIoU(\%)} \\
    \hline
    {MVPNet\cite{jaritz2019multi}} & {2D-3D} & {24.6} & {95} & {64.1} \\
    {PointConv\cite{wu2019pointconv}} & {PointConv} & {21.7} & {307} & {66.6} \\
    {KPConv\cite{thomas2019kpconv}} & {PointConv} & {14.1} & {52} & {68.4} \\
    {DCM-Net\cite{schult2020dualconvmesh}} & {GraphConv} & {0.76} & {151} & {65.8} \\
    {VMNet (\textbf{Ours})} & {Sparse+Graph Conv} & {17.5} & {107} & {\textbf{74.6}} \\
    \hline
    \end{tabular}
    }
\end{table}

\subsection{Results and Analysis} \label{Results}

\noindent\textbf{Quantitative Results.} We present the performance of our approach compared to recent competing approaches on the {ScanNet
benchmark}~\cite{dai2017scannet}
in Table~\ref{table::scannet}. All {the} methods are grouped by the approaches’ inherent convolutional categories {as discussed in Section~\ref{related}.}
As shown in Table~\ref{table::scannet}, {our method leads to} 
a 74.6\% mIoU score, achieving a significant performance gain of 8.8 \% mIoU comparing to the existing {best-performing} graph convolutional {approach, i.e., DCM-Net~\cite{schult2020dualconvmesh}},
and 1.0 \% mIoU comparing to {the} leading sparse convolutional {approach, i.e., MinkowskiNet~\cite{choy20194d}}.
{Our method achieves results comparable to the SOTA 2D-3D method BPNet~\cite{BPNet}, which is a concurrent work on CVPR2021 utilizing both 2D and 3D data while VMNet takes as input only the 3D data.} 
For a fair comparison, the result of OccuSeg \cite{han2020occuseg} is not listed in this table, since it utilizes extra instance labels for training.
{We also evaluate our algorithm on the novel Matterport3D dataset~\cite{Matterport3D} and report the results in Table~\ref{table:matterport}. VMNet achieves overall state-of-the-art results outperforming the previous best method by 1\% in terms of mean class accuracy. Since some methods only report results in one of these two datasets, the listed methods in Tables~\ref{table::scannet} and \ref{table:matterport} are different.}

\begin{figure*}[t]
	\centering
	\includegraphics[width=0.94\linewidth]{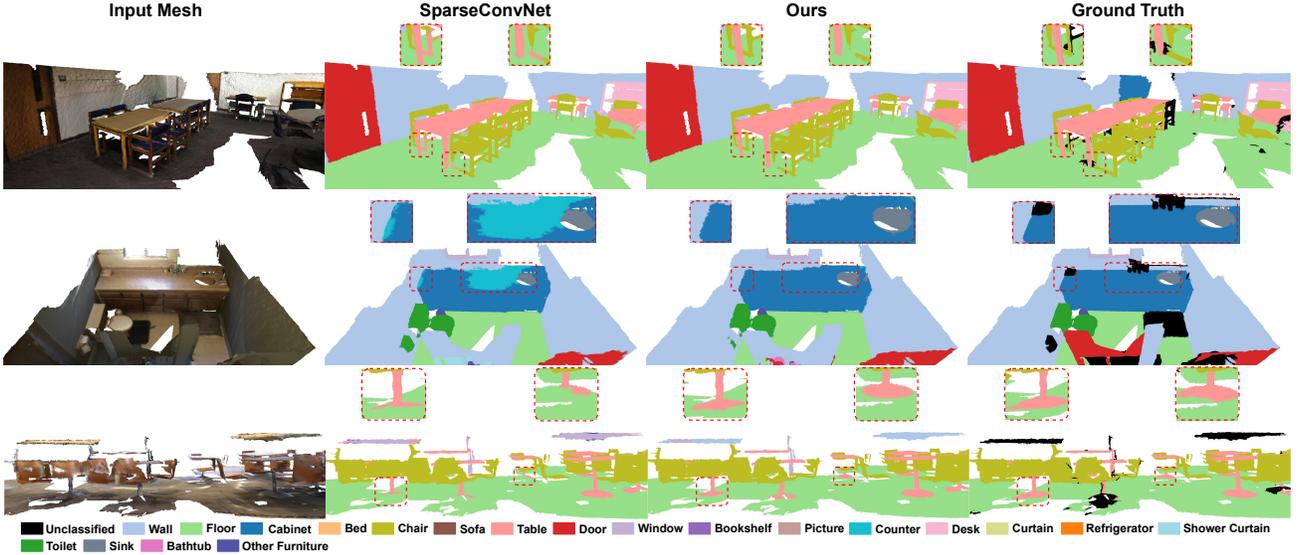}
	\caption{
	\textbf{Qualitative results on ScanNet Val~\cite{dai2017scannet}.} The key parts for comparison are highlighted by dotted red boxes. 
	}
	\label{fig:qualitative}
\end{figure*}

\begin{figure*}[t]
	\centering
	\includegraphics[width=0.94\linewidth]{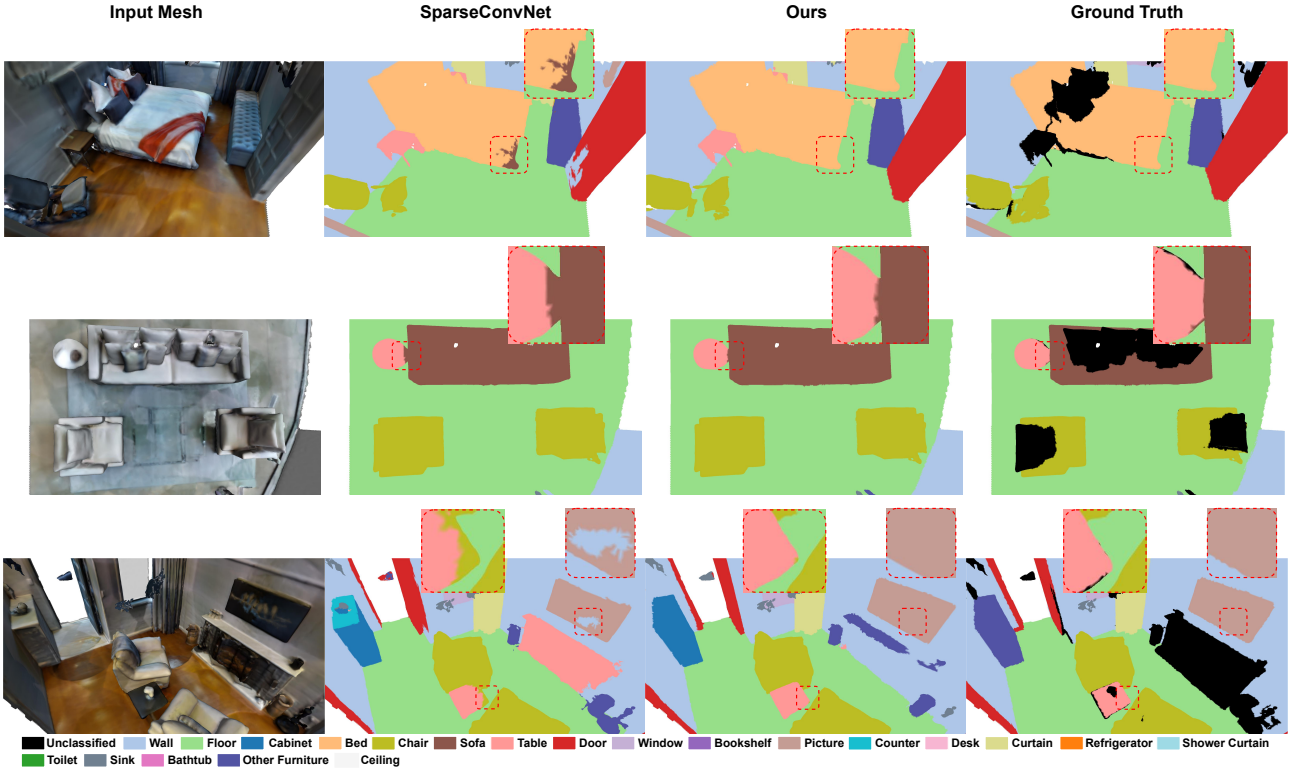}
	\caption{
	{
	\textbf{Qualitative results on Matterport3D Test~\cite{Matterport3D}.} The key parts for comparison are highlighted by dotted red boxes.
    }
	}
	\label{fig:qualitative_m}
\end{figure*}

{
\noindent\textbf{Qualitative Comparison.} Fig.~\ref{fig:qualitative} shows our qualitative results on the ScanNet validation set {and Fig.~\ref{fig:qualitative_m} shows the results on the Matterport3D test set.} 
{Compared to the SOTA sparse voxel-based method SparseConvNet, which operates in the Euclidean domain solely, VMNet generates more distinctive features for close-located objects and better handles complex geometries thanks to the combined Euclidean and geodesic information.} More qualitative results can be found in the supplementary material.
}

{\noindent\textbf{Complexity.} We compare our method with representative SOTA methods of various convolutional categories in terms of their run-time complexity. We randomly select a scene from the ScanNet validation set
and compute the latency results by averaging the inference time of 100 forward passes.

\begin{table}
\caption{\textbf{Ablation study}: \textbf{(Left)} Euclidean and geodesic information; \textbf{(Right)} Network components.}
\label{table::information_component}
\resizebox{0.47\textwidth}{!}{
    \begin{tabular}{r|c}
    \hline
    Information & mIoU(\%) \\
    \hline
    Geo Only & 58.1 \\
    Euc Only & 71.0 \\
    VMNet(Geo+Euc) & \textbf{73.3} \\
    \hline
    \end{tabular}
    \quad
    \begin{tabular}{ccc|c}
    \hline
    Baseline & Intra & Inter & mIoU(\%) \\
    \hline
    \checkmark & & & 70.2 \\
    \checkmark & \checkmark & & 72.1 \\
    \checkmark & \checkmark & \checkmark & \textbf{73.3}\\
    \hline
    \end{tabular}
}
\end{table}

\begin{table}
\caption{\textbf{Ablation study}: \textbf{(Left)} Attentive operators; \textbf{(Right)} Mesh simplification.}
\label{table::attentive_mesh}
\resizebox{0.47\textwidth}{!}{
    \quad
    \quad
    \begin{tabular}{r|c}
    \hline
    Operator & mIoU(\%) \\
    \hline
    Vector Attention & 72.3 \\
    EdgeConv & 72.6 \\
    Scalar Attention & \textbf{73.3}\\
    \hline
    \end{tabular}
    \quad
    \quad
    \begin{tabular}{r|c}
    \hline
    Method & mIoU(\%) \\
    \hline
    VC only & 72.3 \\
    QEM only & 72.9 \\
    VC + QEM & \textbf{73.3} \\
    \hline
    \end{tabular}
    \quad
    \quad    
}
\end{table}

In Table~\ref{table::complexity}, we present the results of two SOTA sparse voxel-based methods, i.e., SparseConvNet~\cite{graham20183d} and MinkowskiNet~\cite{choy20194d}. Although the accuracies of sparse voxel-based methods are not dependent on the implementation of sparse convolution, the latencies of these methods are highly dependent on the implementation. Therefore, we re-implement SparseConvNet and MinkowskiNet using the same version of sparse convolution ({i.e.,} torchsparse~\cite{tang2020searching}) as VMNet for a fair comparison. As shown in the table, VMNet achieves the highest mIoU score with the least number of parameters. It implies that, compared to extracting features in the Euclidean domain alone, combining {the} Euclidean and geodesic information leads to a more effective aggregation of features, even with a simpler network structure. The latency of VMNet is slightly higher than our new implementations of the other two methods. This is caused by the unoptimized projection operations, which are left for future improvement. 

In Table~\ref{table::complexity_sup}, we report more complexity comparisons of our network against other representative methods, including MVPNet\cite{jaritz2019multi}, PointConv\cite{wu2019pointconv}, KPConv\cite{thomas2019kpconv}, and DCM-Net\cite{schult2020dualconvmesh}. While achieving the highest mIoU, VMNet is largely comparable to {these} 
representative methods, in terms of both inference time and parameter size. 
}

\subsection{Ablation Study} \label{Ablation}
In this section, we conduct a number of controlled experiments that demonstrate the effectiveness of building modules in VMNet{,} and also examine some specific decisions in VMNet design. {Since the test set of ScanNet is not available for multiple tests, a}ll experiments are conducted on the validation set, keeping all hyper-parameters the same.

\noindent\textbf{Euclidean and Geodesic Information.}
In Section~\ref{method}, we advocate the combination of Euclidean and geodesic information. To investigate their impacts, we compare VMNet to two baseline networks: {``Euc only'' is a U-Net structure based on sparse convolutions operating on voxels and ``Geo only'' has the same structure but is based on the proposed intra-domain attention layers operating on meshes.}
For a fair comparison, we keep the layer numbers of these baselines the same as the Euclidean branch of VMNet but increase their channel numbers to make sure all the compared methods have similar parameter sizes. As shown in Table~\ref{table::information_component} (Left), VMNet outperforms the two baselines showcasing the benefit of combining information from {the} two domains. 

\noindent\textbf{Network Components.} In Table~\ref{table::information_component} (Right), we evaluate the effectiveness of each component of our method. ``Baseline'' represents the Euclidean branch of VMNet{,} which is a U-Net network built on voxel convolutions. ``Intra'' refers to the intra-domain attentive aggregation module and ``Inter'' refers to the inter-domain attentive fusion module. As shown in the table, by combining the intra-domain attentive aggregation module {with the baseline}, we can improve the performance by 1.9\%. This improvement is brought by the introduction of geodesic information through feature refinement on meshes. From the inter-domain attentive fusion module, we further gain about 1.2\% improvement in performance by adaptive fusion of features from {the} two domains. 

{
\noindent\textbf{Attentive Operators.} In Sections~\ref{intra} and \ref{inter}, we adopt the standard scalar attention~\cite{NIPS2017_3f5ee243} to build the intra-domain attentive aggregation module and the inter-domain attentive fusion module. In Table~\ref{table::attentive_mesh} (Left), we evaluate the influence of different forms of attentive operators in our architecture. ``Scalar Attention'' refers to the operators used in VMNet as presented in Equations~\ref{equation:intra} and \ref{equation:inter}. ``Vector Attention'' represents a variant of Scalar Attention, in which attention weights are not scalars but vectors that can modulate individual feature channels.
It {has been} 
widely adopted in previous attention-based methods operating on 3D point clouds~\cite{wang2019graph,zhao2020point}. Specifically, at layer $k$, the output feature $f_i^{'}$ of vertex $V_i$ with an input feature $f_i$ is computed as (note that the superscripts are ignored here for simplicity):
    \begin{equation}
    \begin{split}
        f_i^{'} = \rho_k(f_i) + \sum_{j \in N_i}\Omega_{ij}\odot\alpha_k(f_j), \\
        \Omega_{ij} = softmax(\gamma_k(\beta_k(\varphi_k(f_i), \psi_k(f_j)))),
        \label{equation:va}
    \end{split}
    \end{equation}
where $N_i$ is the one-ring neighborhood of vertex $V_i$. The functions $\rho_k$, $\alpha_k$, $\varphi_k$, and $\psi_k$ are vertex-wise feature transformations implemented by MLP, $\Omega_{ij}$ is the attention vector, $\odot$ denotes channel-wise multiplication$, \beta_k$ is a relation function (here we implement it by subtraction following~\cite{zhao2020point}), and $\gamma_k$ is a mapping function (implemented by MLP) that produces attention vectors for feature aggregation.

Moreover, we implement a non-attention baseline {built} 
on the popular EdgeConv~\cite{wang2019dynamic}, which is originally proposed to operate on kNN graphs of 3D point clouds. Specifically, at layer $k$, the output feature $f_i^{'}$ of vertex $V_i$ with an input feature $f_i$ is computed as:
    \begin{equation}
    \begin{split}
        f_i^{'} = \sum_{j \in N_i}h_k(f_i || f_j - f_i), 
        \label{equation:dg}
    \end{split}
    \end{equation}
where $N_i$ is the one-ring neighborhood of vertex $V_i$. $h_k$ is a feature mapping function implemented by MLP and $||$ denotes concatenation.

As shown in Table~\ref{table::attentive_mesh} (Left), the scalar attention used in VMNet achieves the best result {and outperforms} 
the non-attention baseline ``EdgeConv'' by 0.7\% and the attentive variant ``Vector Attention'' by 1.0\%. Interestingly, the non-attention baseline ``EdgeConv'' performs slightly better than the attention-based baseline ``Vector Attention''. A possible reason is that ``Vector Attention'' adaptively modulates each individual feature channel and this property appears to be overfitting in our case.
}

\begin{figure}[h]
\centering
\includegraphics[width=0.43\textwidth]{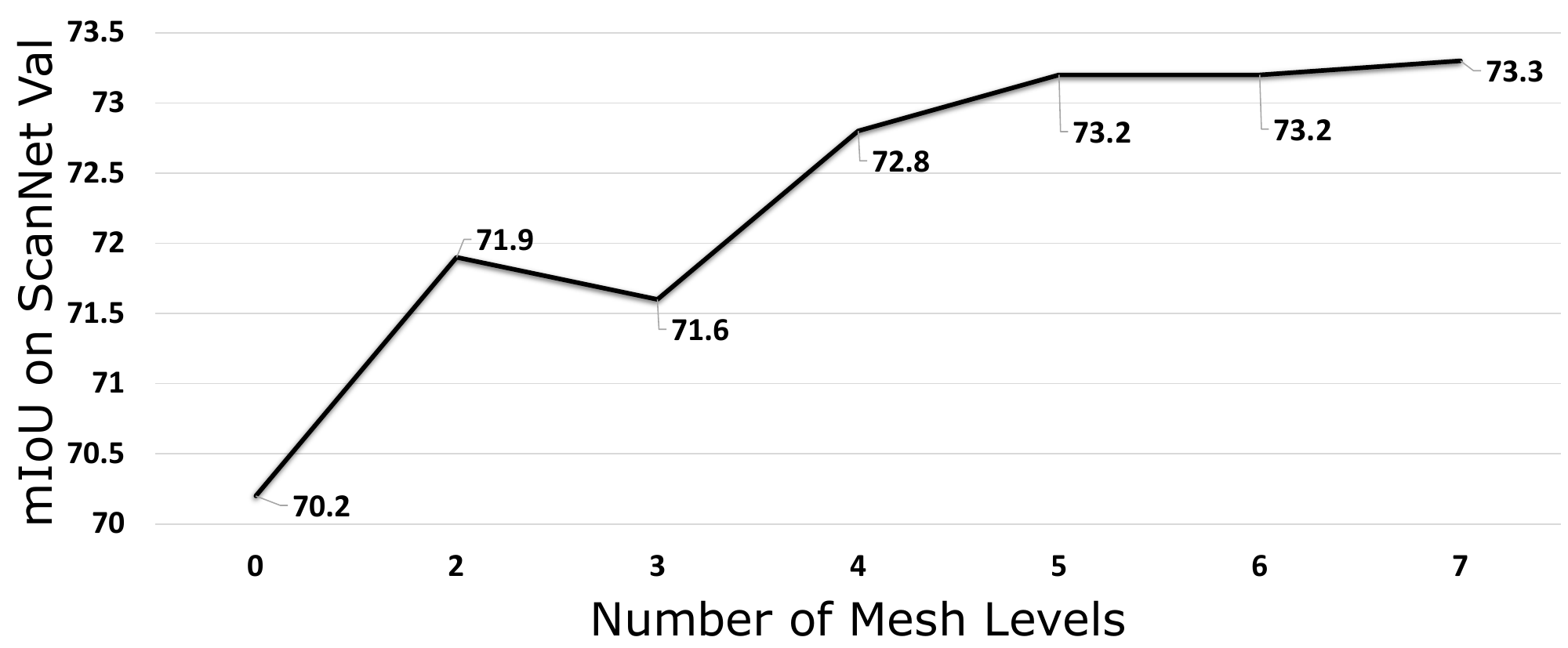}
\caption{{\textbf{Ablation study:} Multi-level feature refinement.}}
\label{fig:multilevel}
\end{figure}

\noindent\textbf{Mesh Simplification.} In Section~\ref{mesh}, we discuss two mesh simplification methods Vertex Clustering (VC) and Quadric Error Metrics (QEM) for multi-level feature learning. 
{We apply the VC method on the first two mesh levels to remove high-frequency signals in noisy areas, and then apply the QEM method on the remaining mesh levels for its better topology-preserving property.}
{To justify our choice, we train three models with the same network definition but performing on different mesh hierarchies, and compare their performances in Table~\ref{table::attentive_mesh} (Right).}
``VC+QEM'' refers to the mesh hierarchy simplified by the combination of the VC and QEM methods as described in Section~\ref{Implementation}. For ``VC only'', at each mesh level $\mathcal{M}^l$, we set the cubical cell lengths of the VC method to the same size as the lengths of voxels in the corresponding voxel level $\mathcal{S}^l$. For ``QEM only'', at each mesh level $\mathcal{M}^l$, the QEM method simplifies the mesh until the vertex number is reduced to 30\% of its preceding mesh level $\mathcal{M}^{l-1}$. As shown in Table~\ref{table::attentive_mesh} (Right), we witness a significant performance gap of 1.0\% between the results of ``VC+QEM'' and ``VC only''. We assume that the more faithful geodesic information provided by meshes simplified through the QEM method leads to {the performance gain}. We also notice that the performance of ``QEM only'' is slightly lower than the one of ``VC+QEM''.
It may {be} caused by the resulting high-frequency noises of  directly applying the QEM method on the original meshes.

\begin{figure}[htp]
	\centering
	\includegraphics[width=0.47\textwidth]{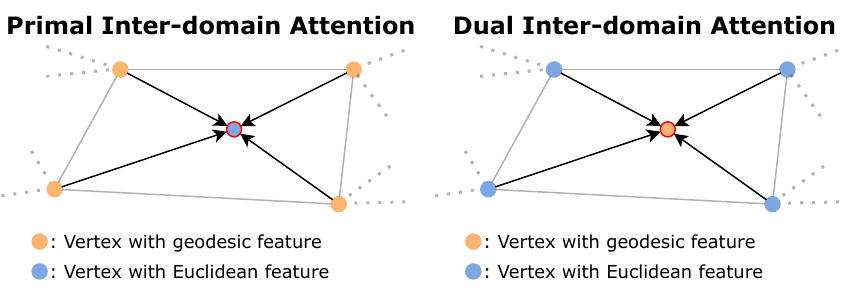}
	\caption{
	{
	\textbf{Illustration of primal and dual inter-domain attention.} \textbf{(Left)} The primal inter-domain attention generates query vectors from the Euclidean features and aggregates the neighboring geodesic features. \textbf{(Right)} The dual inter-domain attention generates query vectors from the geodesic features and aggregates the neighboring Euclidean features. 
	}
	}
	\label{fig:primal-dual}
\end{figure}

{
\noindent\textbf{Multi-level Feature Aggregation and Fusion.}
To measure the effects of {individual geodesic feature refinement levels}, 
we successively add {the} aggregation and fusion modules to the overall architecture. Except {for} the baseline with no geodesic branch, we start with the outermost mesh levels $\mathcal{M}^0 \& \mathcal{M}^1$ to retain one fusion {module} and two aggregation modules. {Next,} along with each added mesh level, one fusion module and one aggregation module are added. The results are presented in Fig.~\ref{fig:multilevel}. We witness that the first four levels bring the most performance gain{,}
indicating the higher importance of finer-level meshes for geometric learning.
}

{
\noindent\textbf{Design Choice of Inter-domain Attention.}
As described in Section~\ref{inter}, we proposed an inter-domain attentive module for adaptive feature fusion. The module takes both the Euclidean features and the geodesic features as input and utilizes the attention mechanism, in which the attention weights are conditioned on features from both {the} domains. To build such an inter-domain attentive module, there are two design choices. 
\begin{figure}[htp]
	\centering
	\includegraphics[width=0.47\textwidth]{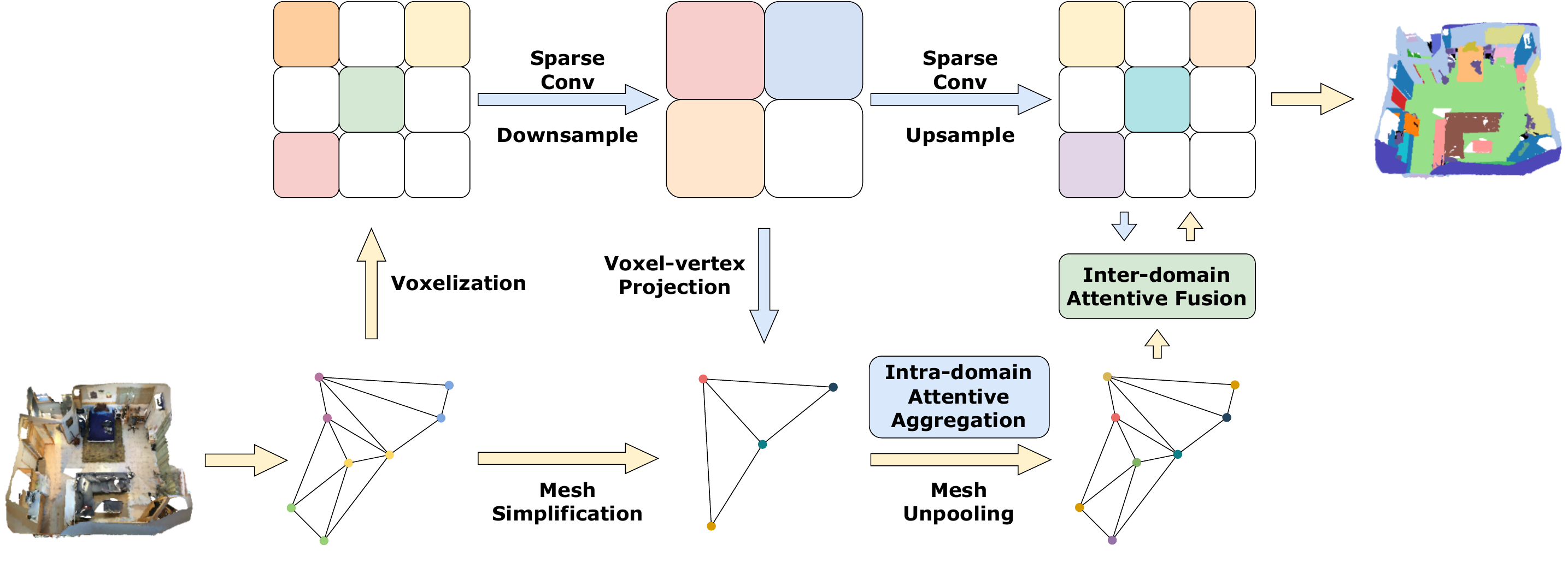}
	\caption{
	{
	\textbf{Network architecture {V}ariant 1.} Sharing the same encoder with VMNet, {V}ariant 1 decodes the features in the Euclidean domain. For simplicity, only two levels of network are shown. 
	}
	}
	\label{fig:variant_1}
\end{figure}
As shown in Fig.~\ref{fig:primal-dual}, we denote the one used in VMNet as the primal inter-domain attention and denote the other one as the dual inter-domain attention. We empirically find that the primal inter-domain attention yields better results than the dual one (73.3\% vs 72.8\% in mIoU on ScanNet Val). It may {be} caused by the different importance of the Euclidean features and the geodesic features in the task of indoor scene 3D semantic segmentation.
}

\begin{figure}[htp]
	\centering
	\includegraphics[width=0.47\textwidth]{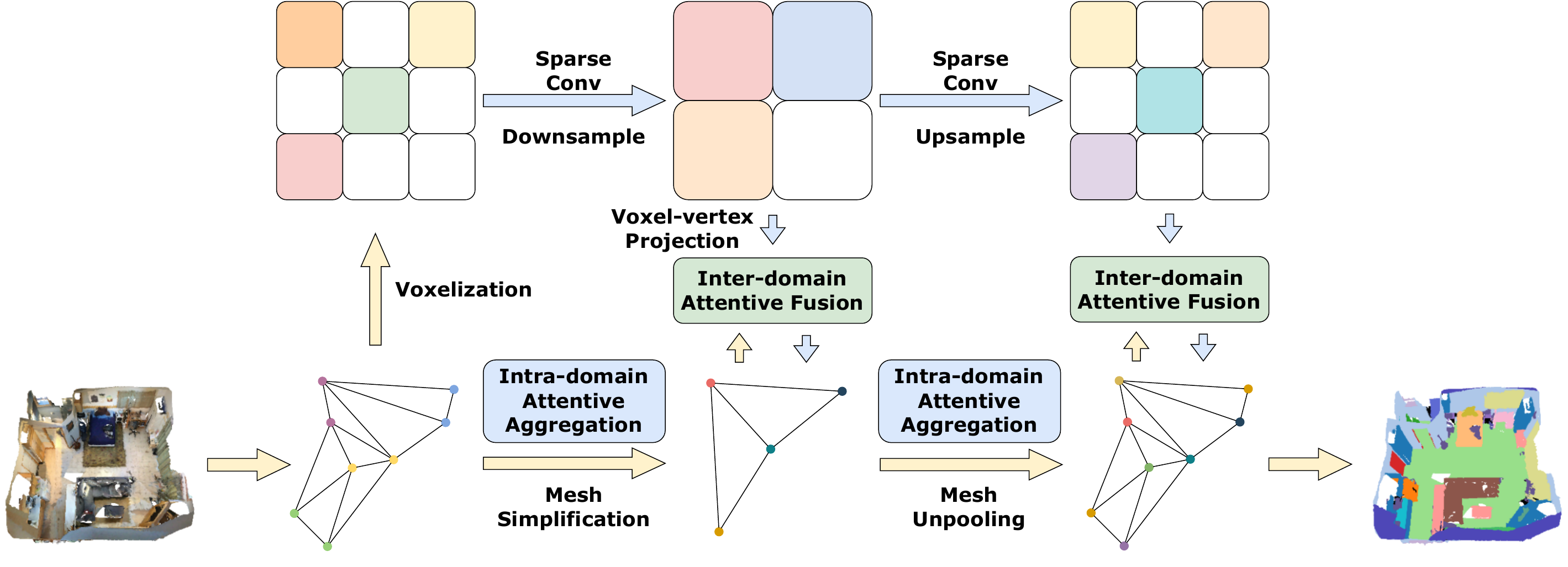}
	\caption{
	{
	\textbf{Network architecture Variant 2.} Sharing the same decoder with VMNet, Variant 2 incorporates both the voxel and mesh representations for contextual feature encoding. For simplicity, only two levels of network are shown.
	}
	}
	\label{fig:variant_2}
\end{figure}

{
\noindent\textbf{Design Choice of Network Architecture.}
As shown in Fig.~\ref{fig:overview}, VMNet encodes contextual information in the Euclidean domain and then decodes the aggregated features in the geodesic domain. To justify our design choice, we construct two variants of VMNet.

Variant 1~(Fig.~\ref{fig:variant_1}) shares the same encoder structure with VMNet. However, in the decoding part, instead of projecting the Euclidean features to the geodesic domain, Variant 1 projects the aggregated and fused geodesic features back to the Euclidean domain, and the per-voxel features on the last voxel level $S^0$ are used for semantic prediction. We compare the performances of Variant 1 and VMNet in Table~\ref{table::variants_na}. While having similar network complexity, Variant 1 performs significantly worse than our proposed VMNet. We {speculate} 
that the problem lies in the projection from vertices to voxels. To better preserve the geodesic information, we apply the QEM method to simplify the meshes resulting in non-uniform vertices. In the projection from vertices to voxels, multiple vertices thus correspond to a single voxel in areas with complex geometries and a voxel in areas with simple geometries may have no corresponding vertices at all. In contrast, projecting from uniform voxels to non-uniform vertices does not have this problem since all the vertices are covered and vertices sharing the same neighboring voxels have distinctive features through trilinear interpolation.

Variant 2~(Fig.~\ref{fig:variant_2}) shares the same decoder structure with VMNet. In its encoding part, we incorporate both the voxel and mesh representations for contextual feature aggregation. As shown in Table~\ref{table::variants_na}, Variant 2 performs slightly better than our proposed VMNet (0.1\% improvement in terms of mIoU) but brings considerable extra complexity. For a better complexity-performance trade-off, we prefer the proposed design choice of VMNet.
}

\begin{table}[h]
    \caption{{\textbf{Ablation study:} Variants of network architecture.}
    }
    \label{table::variants_na}
    \centering
    \resizebox{0.47\textwidth}{!}{
    \begin{tabular}{r|c|c|c}
    \hline
    {Method} & {Params (M)} & {Latency (ms)} & {mIoU(\%)} \\
    \hline
    {Variant 1} & {17.6} & {113} & {72.4} \\
    {Variant 2} & {18.3} & {155} & {\textbf{73.4}} \\
    {VMNet} & {17.5} & {107} & {73.3} \\
    \hline
    \end{tabular}
    }
\end{table}

\section{Conclusion}
In this paper, we have presented a novel 3D deep architecture for semantic segmentation of indoor scenes, named Voxel-Mesh Network (VMNet). {Aiming at addressing the problem of lacking consideration for the geodesic information in voxel-based methods,}
VMNet takes advantages of both the semantic contextual information available in voxels and the geometric surface information available in meshes to perform geodesic-aware 3D semantic segmentation. Extensive experiments show that VMNet achieves state-of-the-art results on the challenging ScanNet and Matterport3D datasets, significantly improving over strong baselines. We hope that our work will inspire further investigation of the idea of combining Euclidean and geodesic information, the development of new intra-domain and inter-domain modules, and the application of geodesic-aware networks to other tasks, such as 3D instance segmentation.

\ifCLASSOPTIONcompsoc
  \section*{Acknowledgments}
\else
  \section*{Acknowledgment}
\fi

This work was partially supported by a Strategic Research Grant from City University of Hong Kong (Project No. 7005729).

\ifCLASSOPTIONcaptionsoff
  \newpage
\fi



%



{
\bibliographystyle{IEEEtran}

}

%

\begin{IEEEbiography}[{\includegraphics[width=1in,height=1.25in,clip,keepaspectratio]{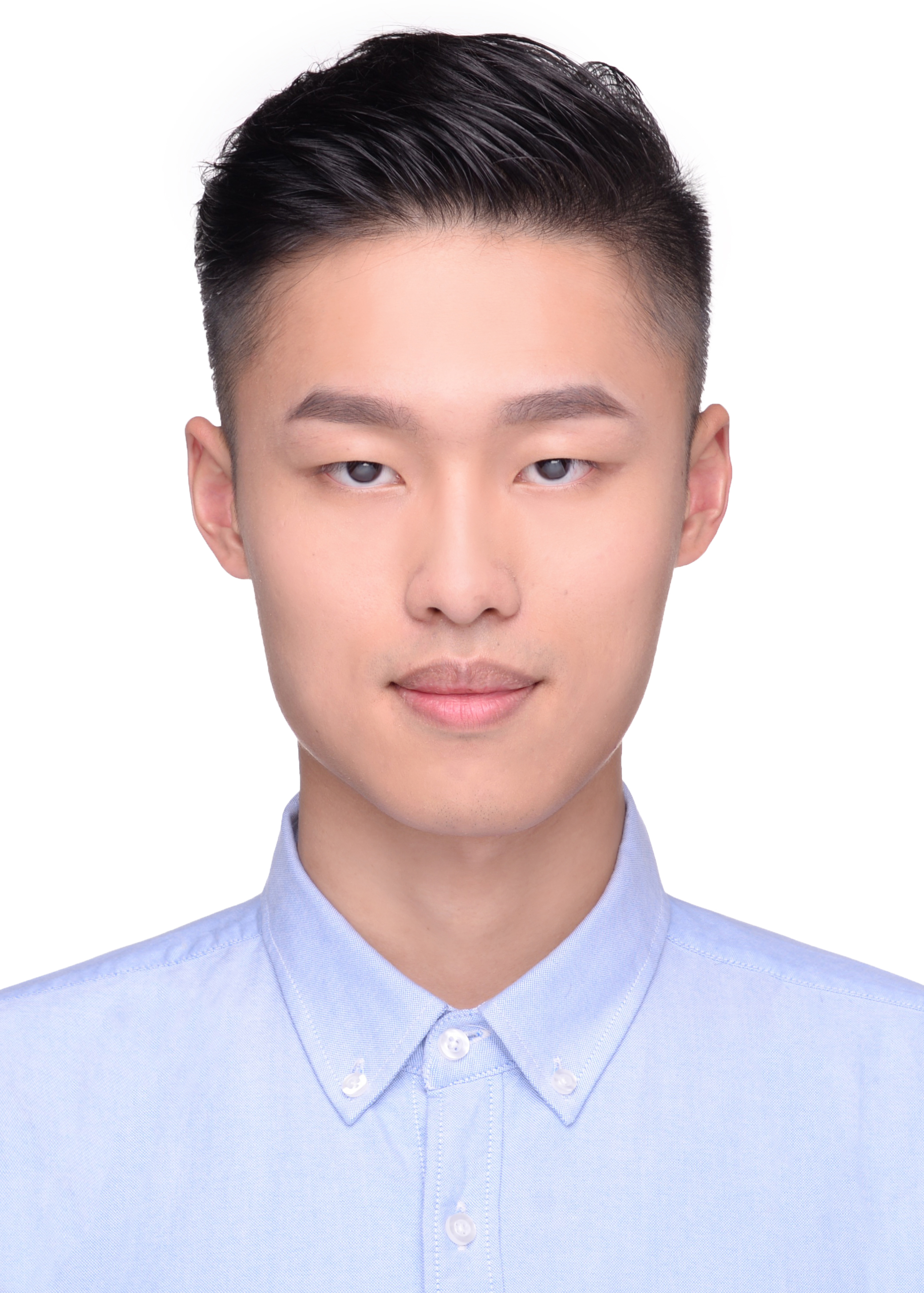}}]{Zeyu Hu}
is a Ph.D. candidate in the Department of Computer Science and Engineering at the Hong Kong University of Science and Technology. Before joining HKUST, he received a Bachelor degree in Automation from University of Science and Technology of China in 2018. His primary research topic is about scene understanding.
\end{IEEEbiography}

\begin{IEEEbiography}[{\includegraphics[width=1in,height=1.25in,clip,keepaspectratio]{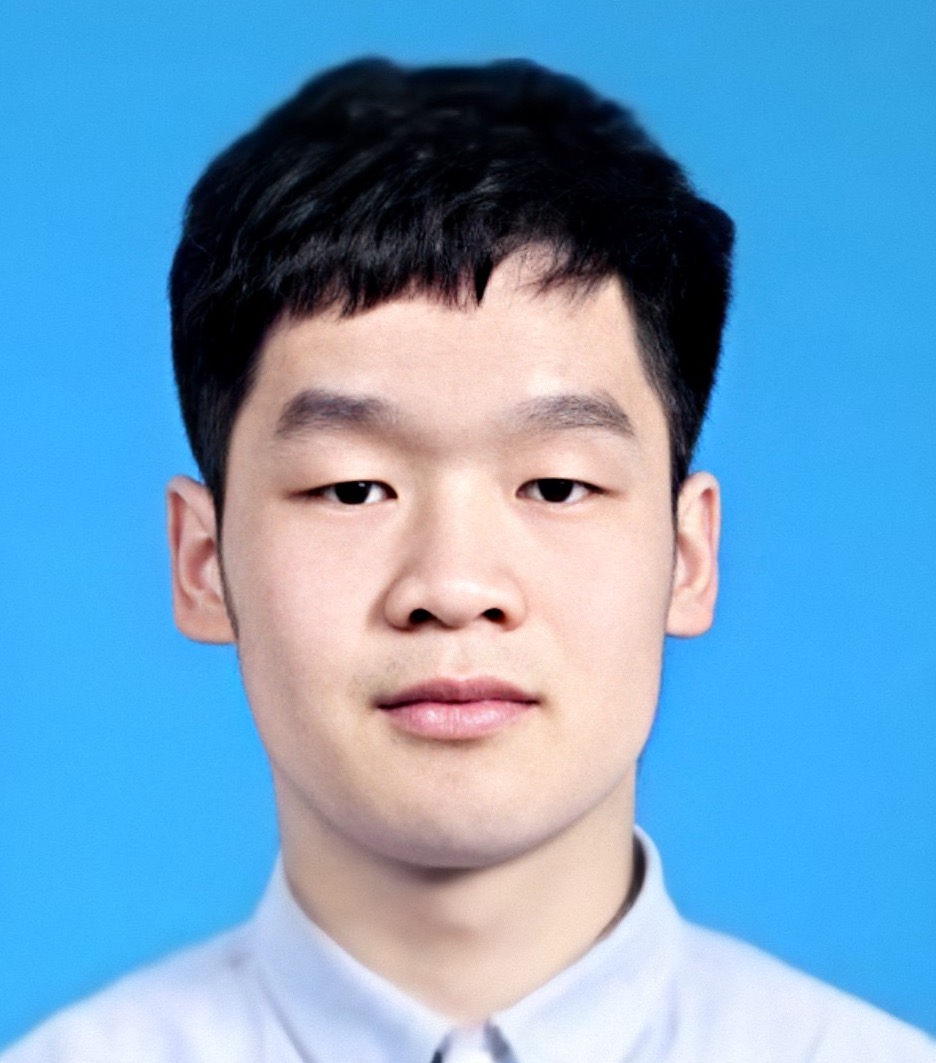}}]{Xuyang Bai}
is a Ph.D. candidate in the Department of Computer Science and Engineering at the Hong Kong University of Science and Technology. Before joining HKUST, he received a Bachelor degree in Electronic Information Science and Technology from Beijing Normal University in 2018. His primary research topic is about point cloud registration and LiDAR perception.
\end{IEEEbiography}

\begin{IEEEbiography}[{\includegraphics[width=1in,height=1.25in,clip,keepaspectratio]{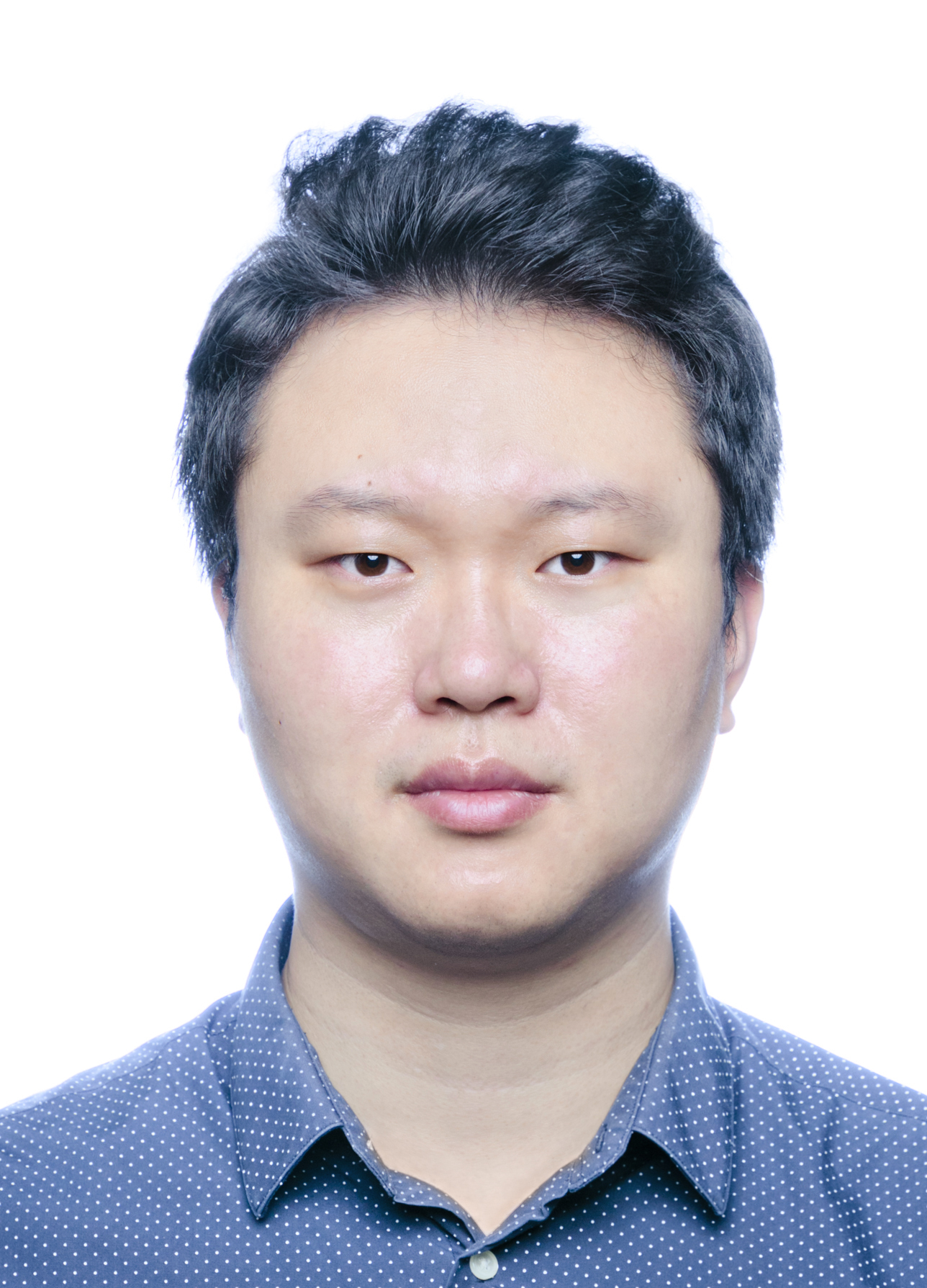}}]{Jiaxiang Shang}
is a Ph.D. candidate in the Department of Computer Science and Engineering at the Hong Kong University of Science and Technology. Before joining HKUST, he had his Bachelor degree in the Department of Computer Science and Engineering at Sun Yat-sen University. His primary research topic is about face reconstruction.
\end{IEEEbiography}

\begin{IEEEbiography}[{\includegraphics[width=1in,height=1.25in,clip,keepaspectratio]{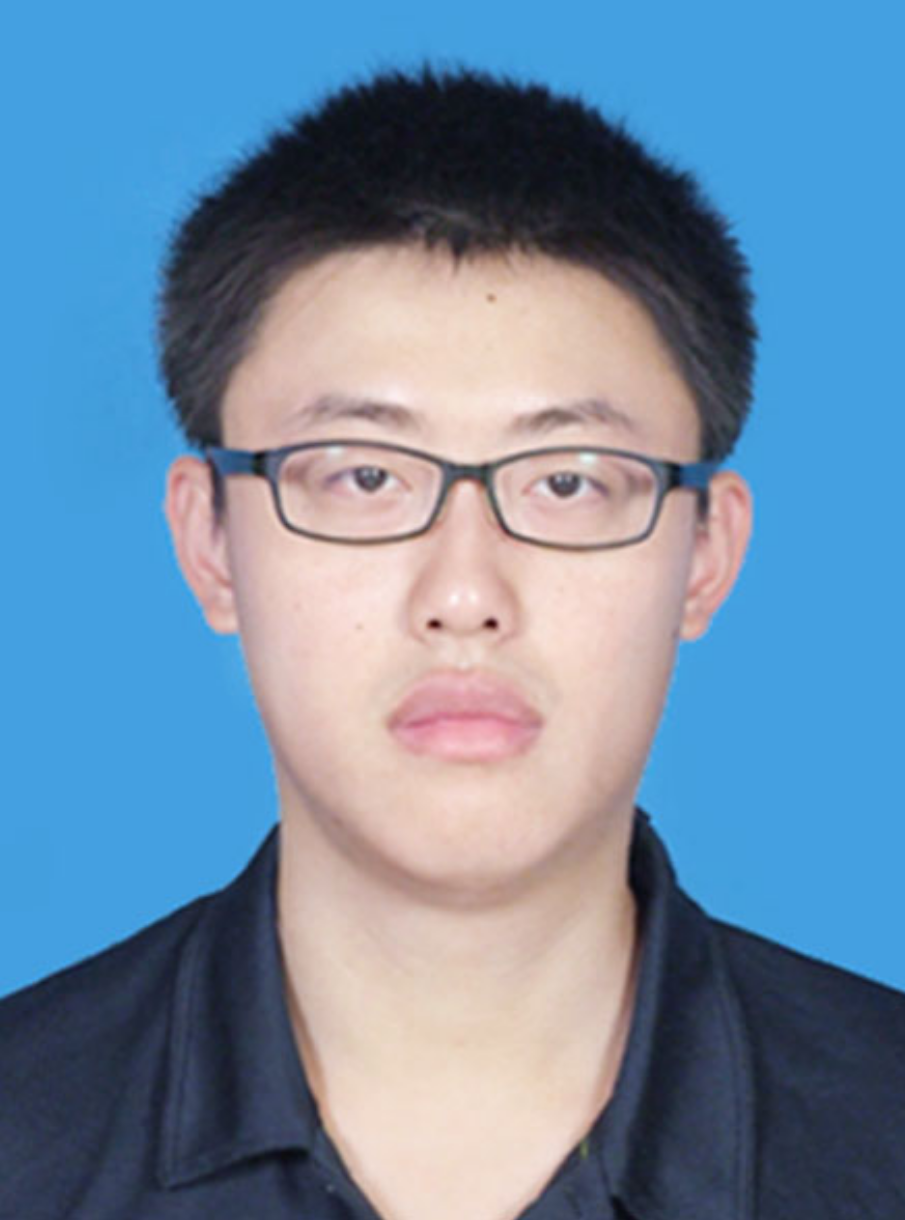}}]{Runze Zhang}
is currently a senior researcher of Tencent, Shenzhen, China. He received the Ph.D. degree from Hong Kong University of Science and Technology, in 2018. He received a Bachelor degree in Intelligence Science and Technology from the Peking University of China in 2013. His research interest is large-scale 3D reconstruction, including Structure-from-Motion, Multi-view Stereo, and other related topics.
\end{IEEEbiography}

\begin{IEEEbiography}[{\includegraphics[width=1in,height=1.25in,clip,keepaspectratio]{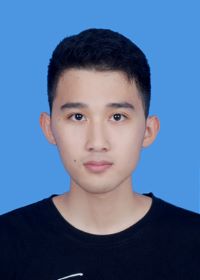}}]{Jiayu Dong}
received the B.S. degree in electronic information science and technology and the M.E. degree in computer science and technology from Sun Yatsen University, Guangzhou, China, in 2017 and 2020, respectively. His current research interests include face recognition and expression recognition.
\end{IEEEbiography}

\begin{IEEEbiography}[{\includegraphics[width=1in,height=1.25in,clip,keepaspectratio]{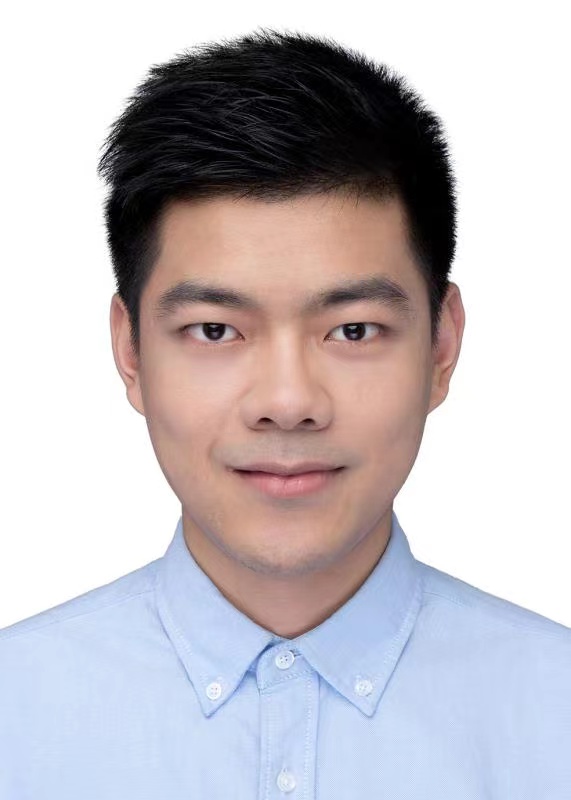}}]{Xin Wang}
is currently a senior researcher of Tencent, Shenzhen, China. He received the Ph.D. degree from the Sorbonne University, in 2017.  His research interests include computer vision, computer graphics and their cross field.
\end{IEEEbiography}

\begin{IEEEbiography}[{\includegraphics[width=1in,height=1.25in,clip,keepaspectratio]{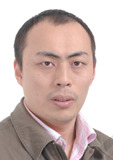}}]{Guangyuan Sun}
received the B.S. degree from Nanjing University, Nanjing, China, in 2004. He is currently a Senior Software Engineer with Tencent Lightspeed \& Quantum Studios, Shenzhen, China. He has been dedicated to various applications of machine learning in game development for many years. His research interests include large-scale learning systems, reinforcement learning, and game theory.
\end{IEEEbiography}

\begin{IEEEbiography}[{\includegraphics[width=1in,height=1.25in,clip,keepaspectratio]{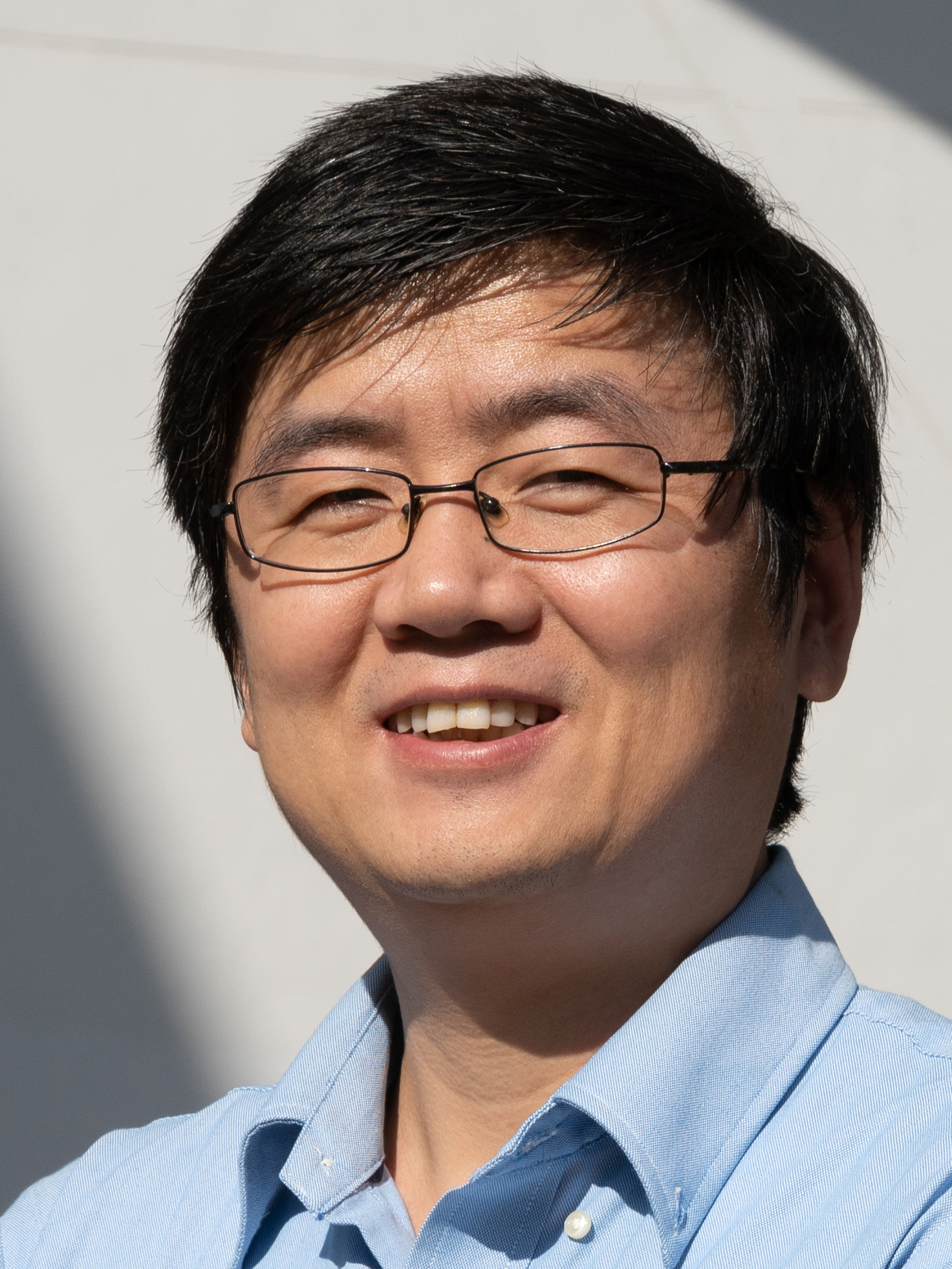}}]{Hongbo Fu}
is a Professor in the School of Creative Media, City University of Hong Kong. Before joining CityU, he had postdoctoral research training at the Imager Lab, University of British Columbia, Canada, and the Department of Computer Graphics, Max-Planck-Institut Informatik, Germany. He received a PhD degree in computer science from the Hong Kong University of Science and Technology in 2007 and a BS degree in information sciences from Peking University, China, in 2002. His primary research interests fall in the fields of computer graphics and human-computer interaction. He has served as an associate editor of The Visual Computer, Computers\&Graphics, and Computer Graphics Forum.
\end{IEEEbiography}

\begin{IEEEbiography}[{\includegraphics[width=1in,height=1.25in,clip,keepaspectratio]{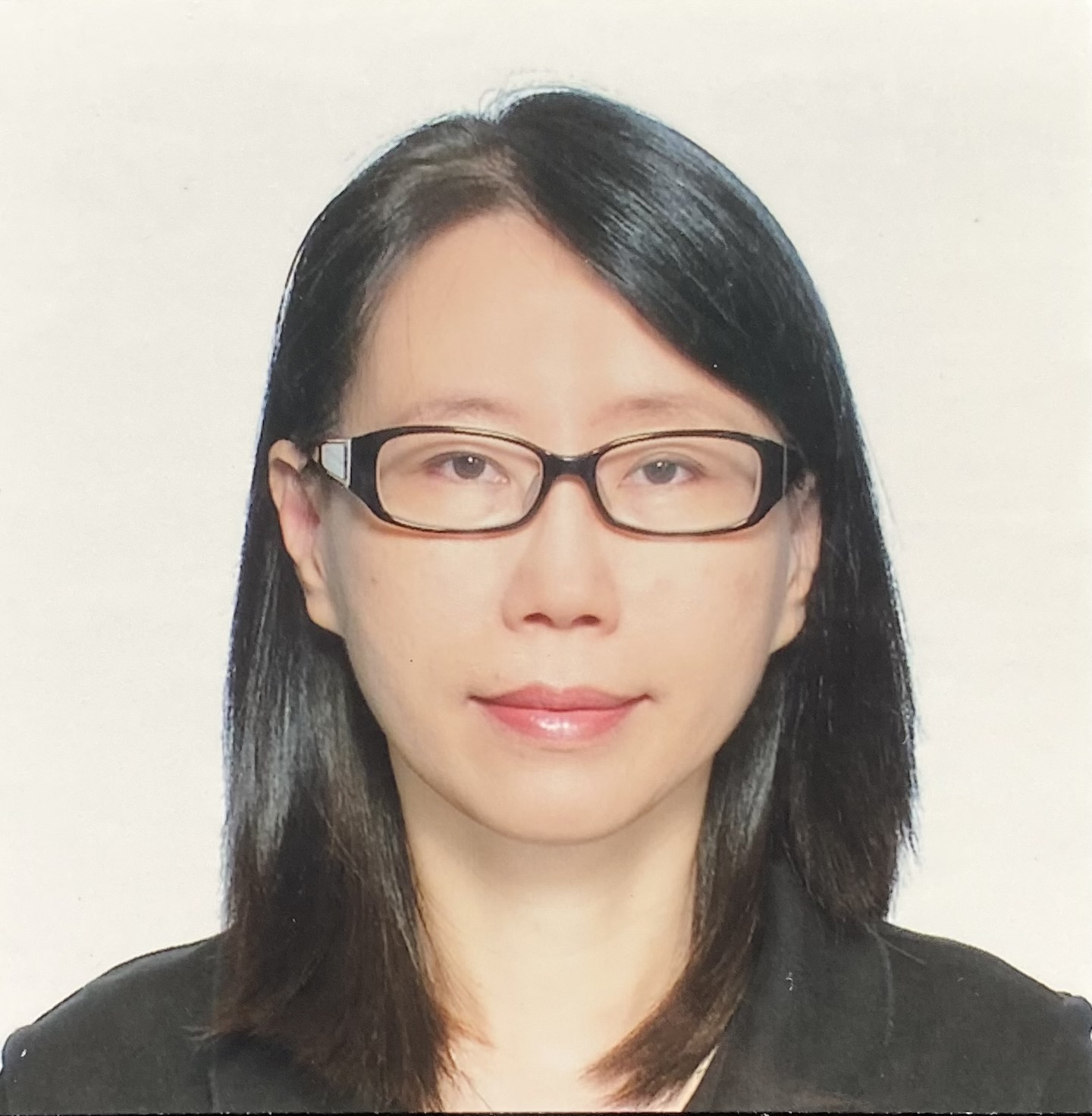}}]{Chiew-Lan Tai}
received the BSc degree in mathematics from the University of Malaya, the MSc degree in computer and information sciences from the National University of Singapore, and the DSc degree in information science from the University of Tokyo. She is currently a professor in the Department of Computer Science and Engineering at the Hong Kong University of Science and Technology. Her research interests include digital geometry processing, computer graphics, and computer vision.
\end{IEEEbiography}







\newpage

\begin{table*}[th]
    \caption{\textbf{Mean intersection over union scores on ScanNet Test~\cite{dai2017scannet}.} This is a detailed version of Table~1 in the main paper.}
    \label{table:scannet_detailed}    
    \centering
    \resizebox{\textwidth}{!}{
        \begin{tabular}{r|c|l|c|c|c|c|c|c|c|c|c|c|c|c|c|c|c|c|c|c|c|c}
        \hline
        Method & mIoU(\%) & Conv Category & \rotatebox[origin=c]{90}{bathtub} & \rotatebox[origin=c]{90}{bed} & \rotatebox[origin=c]{90}{bookshelf} & \rotatebox[origin=c]{90}{cabinet} & \rotatebox[origin=c]{90}{chair} & \rotatebox[origin=c]{90}{counter} & \rotatebox[origin=c]{90}{curtain} & \rotatebox[origin=c]{90}{desk} & \rotatebox[origin=c]{90}{door} & \rotatebox[origin=c]{90}{floor} & \rotatebox[origin=c]{90}{otherfurniture} & \rotatebox[origin=c]{90}{picture} & \rotatebox[origin=c]{90}{refrigerator} & \rotatebox[origin=c]{90}{shower curtain} & \rotatebox[origin=c]{90}{sink} & \rotatebox[origin=c]{90}{sofa} & \rotatebox[origin=c]{90}{table} & \rotatebox[origin=c]{90}{toilet} & \rotatebox[origin=c]{90}{wall} & \rotatebox[origin=c]{90}{window} \\
        \hline\hline
        TangentConv~\cite{tatarchenko2018tangent} & 43.8 & \multirow{8}{3cm}{2D-3D} & 0.437 & 0.646 & 0.474 & 0.369 & 0.645 & 0.353 & 0.258 & 0.282 & 0.279 & 0.918 & 0.298 & 0.147 & 0.283 & 0.294 & 0.487 & 0.562 & 0.427 & 0.619 & 0.633 & 0.352\\
        SurfaceConvPF~\cite{yang2020pfcnn} & 44.2 &  & 0.505 & 0.622 & 0.380 & 0.342 & 0.654 & 0.227 & 0.397 & 0.367 & 0.276 & 0.924 & 0.240 & 0.198 & 0.359 & 0.262 & 0.366 & 0.581 & 0.435 & 0.640 & 0.668 & 0.398\\
        3DMV~\cite{dai20183dmv} & 48.3 &  & 0.484 & 0.538 & 0.643 & 0.424 & 0.606 & 0.310 & 0.574 & 0.433 & 0.378 & 0.796 & 0.301 & 0.214 & 0.537 & 0.208 & 0.472 & 0.507 & 0.413 & 0.693 & 0.602 & 0.539\\
        TextureNet~\cite{huang2019texturenet} & 56.6 & & 0.672 & 0.664 & 0.671 & 0.494 & 0.719 & 0.445 & 0.678 & 0.411 & 0.396 & 0.935 & 0.356 & 0.225 & 0.412 & 0.535 & 0.565 & 0.636 & 0.464 & 0.794 & 0.680 & 0.568\\
        JPBNet~\cite{chiang2019unified} & 63.4 & & 0.614 & 0.778 & 0.667 & 0.633 & 0.825 & 0.420 & 0.804 & 0.467 & 0.561 & 0.951 & 0.494 & 0.291 & 0.566 & 0.458 & 0.579 & 0.764 & 0.559 & 0.838 & 0.814 & 0.598\\
        MVPNet~\cite{jaritz2019multi} & 64.1 & & 0.831 & 0.715 & 0.671 & 0.590 & 0.781 & 0.394 & 0.679 & 0.642 & 0.553 & 0.937 & 0.462 & 0.256 & 0.649 & 0.406 & 0.626 & 0.691 & 0.666 & 0.877 & 0.792 & 0.608\\
        V-MVFusion~\cite{kundu2020virtual} & 74.6 & & 0.771 & 0.819 & 0.848 & 0.702 & 0.865 & 0.397 & 0.899 & 0.699 & 0.664 & 0.948 & 0.588 & 0.330 & 0.746 & 0.851 & 0.764 & 0.796 & 0.704 & 0.935 & 0.866 & 0.728\\
        BPNet~\cite{BPNet} & \textbf{74.9} & & 0.909 & 0.818 & 0.811 & 0.752 & 0.839 & 0.485 & 0.842 & 0.673 & 0.644 & 0.957 & 0.528 & 0.305 & 0.773 & 0.859 & 0.788 & 0.818 & 0.693 & 0.916 & 0.856 & 0.723\\
        \hline
        PointNet++~\cite{qi2017pointnet++} & 33.9 & \multirow{8}{3cm}{PointConv} & 0.584 & 0.478 & 0.458 & 0.256 & 0.360 & 0.250 & 0.247 & 0.278 & 0.261 & 0.677 & 0.183 & 0.117 & 0.212 & 0.145 & 0.364 & 0.346 & 0.232 & 0.548 & 0.523 & 0.252\\
        FCPN~\cite{rethage2018fully}    & 44.7 & & 0.679 & 0.604 & 0.578 & 0.380 & 0.682 & 0.291 & 0.106 & 0.483 & 0.258 & 0.920 & 0.258 & 0.025 & 0.231 & 0.325 & 0.480 & 0.560 & 0.463 & 0.725 & 0.666 & 0.231\\
        PointCNN~\cite{NEURIPS2018_f5f8590c} & 45.8 & & 0.577 & 0.611 & 0.356 & 0.321 & 0.715 & 0.299 & 0.376 & 0.328 & 0.319 & 0.944 & 0.285 & 0.164 & 0.216 & 0.229 & 0.484 & 0.545 & 0.456 & 0.755 & 0.709 & 0.475\\
        DPC~\cite{engelmann2020dilated} & 59.2 & & 0.720 & 0.700 & 0.602 & 0.480 & 0.762 & 0.380 & 0.713 & 0.585 & 0.437 & 0.940 & 0.369 & 0.288 & 0.434 & 0.509 & 0.590 & 0.639 & 0.567 & 0.772 & 0.755 & 0.592\\
        MCCN~\cite{hermosilla2018monte} & 63.3 & & 0.866 & 0.731 & 0.771 & 0.576 & 0.809 & 0.410 & 0.684 & 0.497 & 0.491 & 0.949 & 0.466 & 0.105 & 0.581 & 0.646 & 0.620 & 0.680 & 0.542 & 0.817 & 0.795 & 0.618\\
        PointConv~\cite{wu2019pointconv} & 66.6 & & 0.781 & 0.759 & 0.699 & 0.644 & 0.822 & 0.475 & 0.779 & 0.564 & 0.504 & 0.953 & 0.428 & 0.203 & 0.586 & 0.754 & 0.661 & 0.753 & 0.588 & 0.902 & 0.813 & 0.642\\
        KPConv~\cite{thomas2019kpconv} & 68.4 & & 0.847 & 0.758 & 0.784 & 0.647 & 0.814 & 0.473 & 0.772 & 0.605 & 0.594 & 0.935 & 0.450 & 0.181 & 0.587 & 0.805 & 0.690 & 0.785 & 0.614 & 0.882 & 0.819 & 0.632\\
        JSENet~\cite{hu2020jsenet} & 69.9 & & 0.881 & 0.762 & 0.821 & 0.667 & 0.800 & 0.522 & 0.792 & 0.613 & 0.607 & 0.935 & 0.492 & 0.205 & 0.576 & 0.853 & 0.691 & 0.758 & 0.652 & 0.872 & 0.828 & 0.649\\
        \hline
        SparseConvNet~\cite{graham20183d} & 72.5 & \multirow{2}{3cm}{SparseConv} & 0.647 & 0.821 & 0.846 & 0.721 & 0.869 & 0.533 & 0.754 & 0.603 & 0.614 & 0.955 & 0.572 & 0.325 & 0.710 & 0.870 & 0.724 & 0.823 & 0.628 & 0.934 & 0.865 & 0.683\\
        MinkowskiNet~\cite{choy20194d} & 73.6 & & 0.859 & 0.818 & 0.832 & 0.709 & 0.840 & 0.521 & 0.853 & 0.660 & 0.643 & 0.951 & 0.544 & 0.286 & 0.731 & 0.893 & 0.675 & 0.772 & 0.683 & 0.874 & 0.852 & 0.727\\
        \hline
        SPH3D-GCN~\cite{lei2020spherical} & 61.0 & \multirow{3}{3cm}{GraphConv} & 0.858 & 0.772 & 0.489 & 0.532 & 0.792 & 0.404 & 0.643 & 0.570 & 0.507 & 0.935 & 0.414 & 0.046 & 0.510 & 0.702 & 0.602 & 0.705 & 0.549 & 0.859 & 0.773 & 0.534\\
        HPEIN~\cite{jiang2019hierarchical} & 61.8 & & 0.729 & 0.668 & 0.647 & 0.597 & 0.766 & 0.414 & 0.680 & 0.520 & 0.525 & 0.946 & 0.432 & 0.215 & 0.493 & 0.599 & 0.638 & 0.617 & 0.570 & 0.897 & 0.806 & 0.605\\
        DCM-Net~\cite{schult2020dualconvmesh} & 65.8 & & 0.778 & 0.702 & 0.806 & 0.619 & 0.813 & 0.468 & 0.693 & 0.494 & 0.524 & 0.941 & 0.449 & 0.298 & 0.510 & 0.821 & 0.675 & 0.727 & 0.568 & 0.826 & 0.803 & 0.637\\
        \hline\hline
        VMNet (\textbf{Ours}) & \textbf{74.6} & Sparse+Graph Conv & 0.870 & 0.838 & 0.858 & 0.729 & 0.850 & 0.501 & 0.874 & 0.587 & 0.658 & 0.956 & 0.564 & 0.299 & 0.765 & 0.900 & 0.716 & 0.812 & 0.631 & 0.939 & 0.858 & 0.709\\
        \hline
        \end{tabular}
       }

\end{table*}

\begin{figure*}[t]
	\centering
	\includegraphics[width=\linewidth]{figures/Matterport_TPAMI_supp.pdf}
	\caption{
	\textbf{More qualitative results on Matterport3D Test~\cite{Matterport3D}.} The key parts for comparison are highlighted by dotted red boxes. 
	}
	\label{fig:matterport}
\end{figure*}

\begin{figure*}[t]
	\centering
	\includegraphics[width=\linewidth]{figures/Scannet.pdf}
	\caption{
	\textbf{More qualitative results on ScanNet Val~\cite{dai2017scannet}.} The key parts for comparison are highlighted by dotted red boxes. 
	}
	\label{fig:scannet}
\end{figure*}

\markboth{IEEE TRANSACTIONS ON PATTERN ANALYSIS AND MACHINE INTELLIGENCE, VOL. X, NO. X, MMMMMMM YYYY}%
{Shell \MakeLowercase{\textit{et al.}}: Bare Advanced Demo of IEEEtran.cls for IEEE Computer Society Journals}

\end{document}